\documentclass[conference]{IEEEtran}
\IEEEoverridecommandlockouts
\usepackage{cite}
\usepackage[utf8]{inputenc}
\usepackage{graphicx}
\usepackage{amsmath}
\usepackage{amssymb}
\usepackage[version=4]{mhchem}
\usepackage{siunitx}
\usepackage{longtable,tabularx}
\usepackage{subcaption}
\usepackage{epstopdf}
\usepackage{tabularx}
\usepackage{booktabs} % \**rule table line package
\usepackage{multirow}
\usepackage{arydshln} % dashline package
\usepackage{url}
\usepackage{float}
\usepackage{acronym}
\usepackage{fancyhdr}
\usepackage{threeparttable}
\usepackage{comment}
\usepackage{hyperref} % makes references hyperlinks
\usepackage{textcomp} % Symbol Expansion Package
\usepackage{makecell}
\usepackage{algorithm}
\usepackage{algorithmic}

\usepackage{hyperref} % makes references hyperlinks
\usepackage{textcomp} % Symbol Expansion Package
\def\BibTeX{{\rm B\kern-.05em{\sc i\kern-.025em b}\kern-.08em
    T\kern-.1667em\lower.7ex\hbox{E}\kern-.125emX}}

\renewcommand{\footnoterule}{%
  \kern -3pt
  \hrule width \columnwidth height 1pt
  \kern 2pt
}

\begin{document}

\title{Designing Active Tether-Net Systems for Space Debris Capture with Graph-Learning-Aided Mixed-Combinatorial Optimization}

\makeatletter
\newcommand{\linebreakand}{%
  \end{@IEEEauthorhalign}
  \hfill\mbox{}\par
  \mbox{}\hfill\begin{@IEEEauthorhalign}
}
\makeatother

\author{\IEEEauthorblockN{Feng Liu$^1$ \thanks{$^1$ Ph.D. Candidate, Department of Mechanical and Aerospace Engineering, AIAA Student Member}}
\IEEEauthorblockA{\textit{University at Buffalo}\\
Buffalo, New York, 14260 \\
fliu23@buffalo.edu}
\and
\IEEEauthorblockN{Achira Boonrath$^{1,*}$ \thanks{$^*$Equal contribution}  }
\IEEEauthorblockA{\textit{University at Buffalo}\\
Buffalo, New York, 14260 \\
achirabo@buffalo.edu}
\and
\IEEEauthorblockN{Gishnu Madhu$^{2,*}$\thanks{$^2$ M.S. Student, Department of Computer Science and Engineering}}
\IEEEauthorblockA{\textit{University at Buffalo}\\
Buffalo, New York, 14260 \\
gishnuma@buffalo.edu}
\linebreakand
\IEEEauthorblockN{Eleonora M. Botta$^3$ \thanks{$^3$ Associate Professor, Department of Mechanical and Aerospace Engineering, AIAA member}}
\IEEEauthorblockA{\textit{University at Buffalo}\\
Buffalo, New York, 14260 \\
ebotta@buffalo.edu}
\and
\IEEEauthorblockN{Souma Chowdhury$^4$ \thanks{$^4$ Professor, Department of Mechanical and Aerospace Engineering; Professor (adjunct), Department of Computer Science and Engineering, AIAA Senior Member, Corr. author} 
\thanks{This work is accepted to be presented in the 2026 AIAA Aviation Forum.} 
% \thanks{Copyright \copyright Souma Chowdhury. Personal use of this material is permitted. Permission from AIAA must be obtained for all other uses, in any current or future media, including reprinting/republishing this material for advertising or promotional purposes, creating new collective works, for resale or redistribution to servers or lists, or reuse of any copyrighted component of this work in other works}
}
\IEEEauthorblockA{\textit{University at Buffalo}\\
Buffalo, New York, 14260 \\
soumacho@buffalo.edu}
% \and
% \textit{name of the organization (of Aff.)}\\
}

\maketitle

\newacro{ADR}{Active Debris Removal}
\newacro{CQI}{Capture Quality Index}
\newacro{TSNR}{Tethered Space Space Net Robot}
\newacro{MU}{Maneuverable Unit}
\newacro{RL}{Reinforcement Learning}
\newacro{GNN}{Graph Neural Network}
\newacro{GA}{Genetic Algorithm}
\newacro{DE}{Differential Evolution}
\newacro{PSO}{Particle Swarm Optimization}
\newacro{LHS}{Latin hypercube sampling}
\newacro{MDPSO}{Mixed-Discrete Particle Swarm Optimization}
\newacro{MINLP}{Mixed-Integer Nonlinear Programming}
\newacro{CoM}{Center of Mass}
\newacro{CH}{Convex Hull}
\newacro{MCNLP}{Mixed Combinatorial Nonlinear Programming}
\newacro{MPC}{Model Predictive Control}
\newacro{MSE}{Mean Squared Error}
\newacro{MAE}{Mean Absolute Error}
\newacro{GAT}{graph attention network}
\newacro{GCN}{graph convolutional network}

\thispagestyle{plain}
\pagestyle{plain}
% \begingroup\renewcommand\thefootnote{\textsection}
% \footnotetext{Equal contribution}
% \endgroup

% \thispagestyle{specialfooter}

\begin{abstract}
Active tether-net systems offer a promising solution for capturing large non-cooperative targets with uncertain dynamics, including space debris. This system involves deploying a flexible net from a chaser spacecraft and using maneuverable units (MUs) to manipulate the trajectory of the net. However, concurrent systematic explorations of design and control choices of the tether-net system to understand its full potential remain limited, partly due to the complex, constrained, nonlinear optimization problem that it presents -- one that involves a mixture of continuous, integer and categorical variables, with the latter two arising from net connectivity and component choices, respectively. Classical binary encoding approaches from the combinatorial optimization paradigm do not lend themselves well to solving such Mixed Combinatorial Nonlinear Programmings (MCNLPs) problems in the engineering design domain due to the presence of highly nonlinear, often multimodal, functions as is the case here. On the other hand, integer coding approaches introduce spurious relations among combinations. Given the graph-structured characteristics of the combinatorial space, this paper adopts and extends a new graph-learning-aided optimization approach to solve this MCNLP problem. Here, a Graph Neural Network (GNN) is trained to score (as output) and thereof recommend candidate combinations represented as nodes in a graph, with the continuous variable vector portion of a candidate design given as input. As a result, the MCNLP optimization reduces to an NLP, which can be solved using standard solvers. While this reduction approach is agnostic to the choice of the NLP solver, here a state-of-the-art Particle Swarm Optimization (PSO) algorithm with gradient-based fine-tuning is used as the solver. Demonstrated on the problem of concurrently designing the morphology of the net, choice of mass and thrusters in the MUs and aiming points used by the controller of the tether-net system, the GNN-based recommender is shown to provide significantly faster convergence to similar optimal solutions, compared to direct solution of the MCNLP problem.

\end{abstract}

% \newacro{ADR}{Active Debris Removal}
% \newacro{RL}{Reinforcement Learning}


\section{Introduction}
Space debris poses an increasing risk to space missions in Earth's orbit \cite{anzmeadorIridiumCosmos}. Approaches to solve this problem fall under \ac{ADR}, which can be achieved with various methods, among which tether-net systems are particularly promising for their ability to capture uncooperative and tumbling large-scale debris while maintaining the chaser satellite at a safe distance \cite{shanADRreview, boonrath2025robustness, hou2021dynamic, huang2023contact}. 
% A passive tether-net system consists of a net with 4 or more corner masses that is initially stored on the chaser satellite, a closing mechanism on the net, and a main tether that connects the net to the chaser. 
For robustness of space debris capture missions (e.g., improve the capture success rate), the use of an active and robotic tether-net system has been proposed in more recent years \cite{approach, ZhaoYakun2020ISSM, liu2023learning, boonrath2024learning, zhu2024multi}. Such systems have multiple \acp{MU} attached to the corners or edges of the net, whose thrusters can transport the net towards the debris. After the debris is captured, the chaser initiates a deorbiting or relocation process to transport it to a graveyard orbit. 
% after launch from the chaser. %, so the trajectories of the net can be controlled more subtly.
%
Methods for controlling the \acp{MU} that have been considered in the literature include closed-loop and open-loop approaches \cite{liu2025surrogate,approach, zhu2024multi, boonrath2024learning}. 
% Closed-loop control generally enables more precise maneuvering of the \acp{MU}, making it the preferred method for these tasks. 
For example, Meng et al. proposed a dual-loop control scheme for the deployment of a robotic tether-net system \cite{approach}. Zhu et al. developed a feedback controller that would allow a robotic tether-net system to capture multiple small pieces of space debris located in close proximity \cite{zhu2024multi}. Boonrath et al. proposed a robotic net system with a proportional-integral-derivative thrust controller that is guided by a reinforcement learning policy for debris-capture tasks \cite{boonrath2024learning}. To enable such autonomous control processes, it is important to carefully consider the structure, components (that support autonomy, e.g., sensors/actuators) and control choices, instead of simply adopting traditional (baseline) physical form for the net and corner maneuverable units (MUs). This paper presents a computational framework that allows efficiently searching over the complex mixed continuous-combinatorial design spaces presented by such joint consideration of structure, components and control choices; secondly, it provides a proof of concept for how graph neural nets can be used to systematically (as opposed to ad hoc) and meaningfully (as opposed to indexing) abstract the space of combinatorial options; this could be useful for automated guided design of other autonomous spacecraft systems as well. 
% However, in most of these works, relatively little attention is paid to the optimal guidance (i.e., trajectory planning) aspect of deploying the maneuverable net system. 
%Given this lineage, a closed-loop control approach will be employed for this work as well.

Simultaneously optimizing the physical design and the controlled trajectory of such systems is a challenging task. First, the hardware selections for the tether-net systems may inherently involve fixed-parameter groupings rather than independent variables only. In the example of selecting thrusters, key parameters such as the thrust saturation limit, specific impulse, and dry mass are predefined by the manufacturers and are fixed for each thruster type. In recent years, more methods have been proposed to solve this type of \ac{MCNLP} problem in design or optimal control problems. For example, to efficiently address the combinatorial aspects of the satellite layout problem, Xia et al. \cite{XIA2025103415} formulated it as a mixed-integer program and used a bi-level optimization approach that separates discrete and continuous variables. Ringkamp et al. \cite{RingkampMaik2017Ottt} implemented the concept of time transformation, which converts discrete, integer-controlled time intervals into continuous time functions, thereby smoothing out the integer-control problem. Though the above methods prove to be capable of handling the \ac{MCNLP} problems in some engineering problems, they rely on direct computing of the objective functions for every candidate combination, which is computationally expensive if the simulation involved in the objective functions is computationally expensive; furthermore, the relations of the candidates in the combinatorial space in those methods are not depicted illustratively.

On the other hand, graphs have been proven to be useful to abstract the combinatorial space in the field of classical and contemporary planning problems (and some design problems) that are formed as combinatorial optimization problems, and the approximate solutions can be learned using \acp{GNN} \cite{GNN_survey,vaswani2017attention}. Such problems include vehicle routing \cite{ref2}, multi-robot task allocation \cite{ref3,paul2024learning}, and molecular design \cite{ZhangShiqiang2024Aomd}. In our previous research, \acp{GNN} have successfully been used to optimize both the physical design and the controlled trajectory of the tether-net system in a challenging environment \cite{gnn_reco_idetc}: a \ac{GCN} \cite{ref5} was utilized to efficiently recommend the best combinatorial components (e.g., selection of the thruster) for a given vector of non-combinatorial variables (e.g., control parameters). More \ac{GNN} structures remain to be explored for the tether-net co-design problem, such as \acp{GAT} \cite{ref6} and capsule networks \cite{ref7}.

Motivated by the above related research, we propose to use an updated \ac{GNN}-aided \ac{PSO} framework from our parallel research to minimize the fuel expended in the deployment maneuver by the tether-net system while ensuring successful capture; this is done by optimizing its physical design and the employed control. Compared to our previous research \cite{gnn_reco_idetc}, the main differences are as follows: \textbf{1)} the structure of the \ac{GNN} has been switched from a \ac{GNN} to an encoder-decoder architecture comprising a \ac{GNN} encoder and an MLP decoder. This architecture is referred to as Edge Flow Graph Network. \textbf{2)} The graph abstraction has been enhanced to represent a directed graph -- with edges encoding pairwise objective differences between nodes --, which is the ground truth vector to learn. While in earlier work node-level scores were learned, the current formulation focuses on edge-level signals. This abstraction enables the effective use of edge relationships and allows the direct encoding of the difference between fitness scores, which inherently represent the problem's mixed combinatorial nature, onto the graph. \textbf{3)} The constraints of the problem have been formulated into penalty functions. \textbf{4)} The design space of the tether-net system has been increased to include more continuous and combinatorial variables. \textbf{5)} The final (i.e., end-of-deployment) desired velocity of each \ac{MU} is now determined by the optimization framework, which, while increasing the complexity of the control framework, also widens the range of possible \ac{MU} maneuvers.
%\textbf{5)} The controller of the \acp{MU} now controls the \ac{MU} velocities of reaching the aiming point, which increases the complexity of the control system, but also enhances the maneuverability of the \acp{MU}.

The key contributions of the paper can be summarized as follows: \textbf{1) Specialized \ac{GNN} Loss Function:} We design a specialized loss function to train the \ac{GNN} and a corresponding metric (e.g., sign accuracy) to evaluate the performance of the trained \ac{GNN}, enabling robust recommendation of combinatorial choices given the continuous portion of a candidate design. \textbf{2) Optimization Framework:} We develop a framework for joint structure-component-control optimization with this embedded \ac{GNN} to improve the maneuvering efficiency of the active tether-net system, specifically for minimizing the fuel cost of the maneuver, while preserving successful capture of the debris. In this work, we use \ac{PSO} as the optimizer, with its results post-processed by sequential quadratic programming for optimality assurances; however, the proposed framework is also compatible with other population-based or gradient-based algorithms. \textbf{3) Tether-Net Simulation:} We updated our active tether-net system with a \ac{MPC} system (instead of the earlier PID system) serving as the low-level controller that tracks a minimum-energy reference path to achieve the aiming points (high level control decisions, that is optimized jointly with design).

% \textbf{3) Optimizing Tether-Net Design with a Level of Generalization: } We sampled 10 strategically-selected scenarios for training the \ac{GNN} and operating the optimization, which leads to the optimized designs with a level of generalization capability w.r.t. different target states.

The subsequent sections of this paper are arranged as follows: Section \ref{sec:tns} introduces the active tether-net system, its components, and the controller for the \acp{MU}. Section \ref{sec:opt} introduces the optimization framework and formulations, along with the embedded \ac{GNN} model. Section \ref{sec:results} discusses the results of the optimization in the test scenario. Section \ref{sec:con} then concludes the paper and discusses the future work for this research.

\section{Tether-Net Design Optimization}\label{sec:tns}
\subsection{System Components, Mission Procedure, and Evaluation of Capture Success}
% \begin{figure}[ht!][ht]
%   \centering
%   % ---------- Subfigure (a) ----------
%   \begin{subfigure}[t]{0.7\linewidth}
%     \centering
%     \includegraphics[width=\linewidth]{Images/tether_net_all.pdf}
%     \caption{Diagram of the tether-net system. The crosses are examples aiming points of the \acp{MU} for illustrative purposes.}
%     \label{fig:tethernet_aim}
%   \end{subfigure}
%   \hfill
%   % ---------- Subfigure (b) ----------
%   \begin{subfigure}[t]{0.9\linewidth}
%     \centering
%     \includegraphics[width=\linewidth]{Images/opti_vars_illustrative.pdf}
%     \caption{Illustrations of the selected optimization variables, with $K_{\text{cls}}$ at lower bound in each $N_k$ case as the example.}
%     \label{fig:close_node}
%   \end{subfigure}
  
%   % ---------- Combined caption ----------
%   \caption{Illustrations of the tether-net system}
%   \label{fig:combined_tethernet}
%   % \vspace{-8pt}
% \end{figure}

The net-based \ac{ADR} system of interest for this work includes a robotic net with a closing mechanism interlaced with its perimeter, a main tether connecting the net to a chaser vehicle, and an autonomous chaser vehicle. The chaser is a rigid body that is assumed to be equipped with a winch, all the required sub-systems (including propulsion), and a sensing system that enables estimating the relative location, velocity, and angular states of the target debris. For the net itself, at the corners of the square-shaped structure, \acp{MU} and the closing mechanism threads are attached; in simulation, the net structure is modeled as a collection of particles linked together via spring-damper elements that cannot withstand compression. Here, each \ac{MU} represents a nanosatellite-sized spacecraft that can be controlled using small thrusters and inertial measurement units for navigation. Figure \ref{fig:tethernet_aim} displays the components of the system and the reference frames used in this work, including the inertial reference frame $\mathcal{O}=\{O,\mathbf{\hat{i}},\mathbf{\hat{j}},\mathbf{\hat{k}}\}$ -- fixed to the initial location of the net's corner knot adjacent to \ac{MU} 1 -- and the target body-fixed frame $\mathcal{D}=\{D,\mathbf{\hat{d}}_x,\mathbf{\hat{d}}_y, \mathbf{\hat{d}}_z\}$ -- fixed to the target's \ac{CoM}. The target debris to be captured -- with its size and mass information listed in Table \ref{tab:phy_info} -- is a rigid-body in-simulation that is modeled after the upper stage of the Zenit-2 rocket. Table \ref{tab:phy_info} also lists the chaser spacecraft's properties. The physical properties of the net (except for the axial damping ratio of the net threads, set at 0.106) will be determined through the optimization process defined later in this work. 
%are selected as optimization variables for this work.

% The target is initialized at position $\textbf{r}_t = (9.0, 9.0, -60.0)$ m with respect to $\mathcal{O}$. Its initial orientation is specified by Euler angles $(60.0^\circ, 40.0^\circ, 0.0^\circ)$ and the angular velocity is$ \omega_t = (10.0, 30.0, 10.0)$ deg/s, expressed in $\mathcal{D}$. This scenario has the farthest distance from the chaser with highest angular velocity, which is considered as one of the most challenging scenarios in the target state space from our previous research~\cite{boonrath2024learning, liu2025surrogate}.
%other mission-related quantities, modified from the previous work \cite{liu2025surrogate}. 
% Net properties not listed in Table \ref{tab:phy_info} %-- including thread thickness and Young's Modulus -- 
% are selected as optimization variables for this work.

% \begin{table}[h]
%   \centering
%   \begin{threeparttable}
%   \caption{Tether-Net Simulation component Information}
%   \label{tab:state_action}
%   \begin{tabular}{lllll}
%     \toprule
%     & Parameter & Mass & Diameter & Length\\
%     \midrule
%     \multirow{3}{*}{Debris} & Value & 9000 & 3.9 & 11.0\\
%                 & Unit & kg & m & m \\
%     \bottomrule 
%   \end{tabular}
%   \begin{tablenotes}
%     \item[*] $N$ is the number of \acp{MU}.
%   \end{tablenotes}
%   \end{threeparttable}
%   % \vspace{-15pt}
% \end{table}

\begin{table}[h]
  \centering
  \begin{threeparttable}
  \caption{Debris and Chaser Component Setting}
  \label{tab:phy_info}
  \begin{tabular}{llll}
    \toprule
    & Parameter & Value & Unit\\
    \midrule
    \multirow{5}{*}{Debris} & Mass & 9000 & kg\\
                & Diameter & 3.9 & m \\
                & Length & 11.0 & m\\
                & Volume & 125.3 & m$^3$\\
                & Surface Area & 159.9 & m$^2$\\
    \midrule
    \multirow{2}{*}{Chaser} & Mass & 1600 & kg\\
                & Side length & 1.5 & m \\
    % \midrule
    % \multirow{1}{*}{MU} & Mass (without thruster) & 2.5 & kg \\
    % \midrule
    % \multirow{3}{*}{Net} & Side Length & 22.0 & m \\
    %           %& Mesh Length & 1.0 & $m$/$s$\\÷
    %           % & Thread Radius & 0.0011 & m\\
    %           % & Thread Density & 1390 & kg/m$^3$\\
    %           % & Young's Modulus & 70 & GPa\\
    %           & Axial Damping Ratio & 0.106 & \\
              % & Corner Thread Length & 1.0 & m\\
              % & Corner Thread Radius & 0.0007 & m\\
              
    \bottomrule 
  \end{tabular}
  % \begin{tablenotes}
  %   \item[*] $N$ is the number of \acp{MU}.
  % \end{tablenotes}
  \end{threeparttable}
  % \vspace{-15pt}
\end{table}

The thrusters on the \acp{MU} are switched on at the beginning of the mission to launch the net from the chaser satellite and maneuver it towards the target. 
After 25 seconds, the thrust is deactivated, and the net is closed by shortening and then locking the closing mechanism threads, interlaced around the net. The number of locked segments of the closing mechanism, $N_\text{L}$, and the \ac{CQI} are selected as metrics to determine whether a capture is successful or not. The \ac{CQI} is defined as:
%The number of locked segments of the closing mechanism, $N_\text{L}$, has a maximum of 12, with both $N_\text{L}$ and \ac{CQI} selected to determine whether a capture is successful or not. The \ac{CQI} is defined as:
\begin{equation}
 J = 0.1\frac{|V_n-V_D|}{V_D}+0.1\frac{|S_n-S_D|}{S_D} +0.8\frac{|q_n|}{L_c}
 \label{eq:cqiSafe}
\end{equation}
In the equation, variables are defined as follows: the \ac{CH} volume of the net is $V_n$, the debris's volume is $V_D$, the \ac{CH} surface area of the net is $S_n$, the surface area of the debris is $S_D$, the distance between the debris \ac{CoM} and the net \ac{CoM} is $q_n$, and the minimum distance from the debris's \ac{CoM} to its surface is $L_c$. The value of \ac{CQI} quantifies the similarity between the shape of the debris and the net, as well as the distance between their \acp{CoM}. 
For this work, the \ac{CQI} value at the end of the mission (i.e., 10 s after net closing activation) is indicated with $I^*_{\text{CQI}}$, and a capture simulation with $I^*_{\text{CQI}} \leq 2.5$ and $N_{\text{L}} = 12$ (i.e., the maximum number of locked segments) is defined as successful. In this work, all the simulations are executed on a Python-based simulator, where Google JAX's \texttt{jax.numpy.array} data structure is employed to enable accelerated numerical computation performance \cite{liu2025surrogate}. Interested readers may refer to previous works by the authors for more information on the system's in-simulation modeling \cite{botta2016simulation, botta2019simulation, liu2025surrogate}.

\begin{figure*}[ht!]
  \centering
  % ---------- Subfigure (a) ----------
  \begin{subfigure}[t]{0.5\linewidth}
    \centering
    \includegraphics[width=\linewidth]{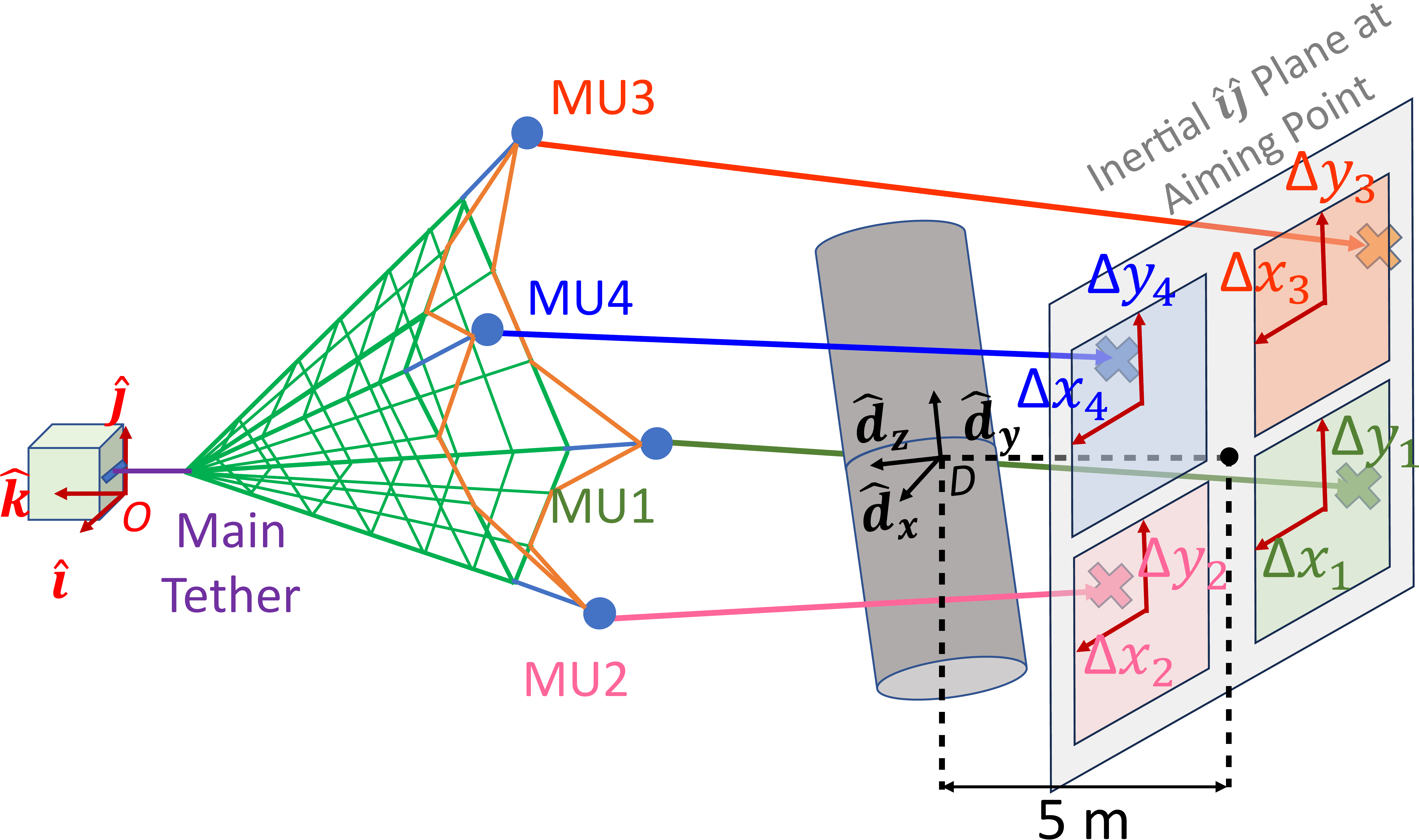}
    \caption{Diagram of the tether-net system. The crosses are examples aiming points of the \acp{MU} for illustrative purposes.}
    \label{fig:tethernet_aim}
  \end{subfigure}
  % \hfill
  \hfill
  % ---------- Subfigure (b) ----------
  \begin{subfigure}[t]{0.8\linewidth}
    \centering
    \includegraphics[width=\linewidth]{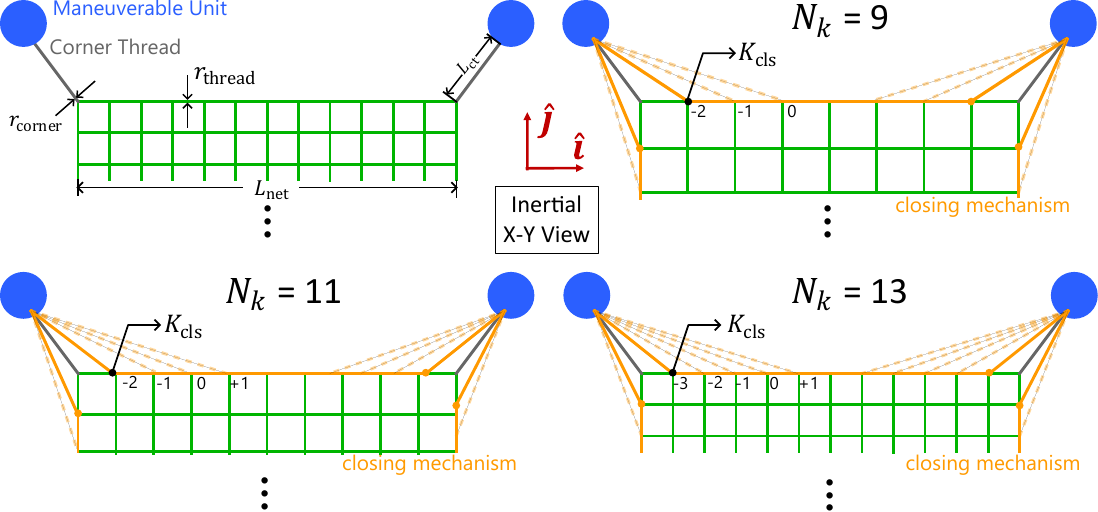}
    \caption{Illustrations of the selected optimization variables, with $K_{\text{cls}}$ at lower bound in each $N_k$ case as the example.}
    \label{fig:close_node}
  \end{subfigure}
  
  % ---------- Combined caption ----------
  \caption{Illustrations of the tether-net system}
  \label{fig:combined_tethernet}
  % \vspace{-8pt}
\end{figure*}

\subsection{Maneuvering Control Process}
In this work, we aim to employ the proposed optimization framework to determine the optimal controlled trajectory of the \acp{MU} and the physical design of the debris removal system. Here, the control formulation for the net deployment is established, based on \ac{MU} control frameworks previously employed by Boonrath et al. \cite{boonrath2024learning, boonrath2025mission}. The control forces for the net deployment (activating and deactivating at instances mentioned in Section~\ref{sec:tns}) are determined through a path-tracking \ac{MPC} controller.

For each \ac{MU}, the tracked reference path is based on a \textit{minimum-energy solution for a reference \ac{MU} model without the net attached} to travel from a set initial state -- that is, a set combination of position and velocity -- to a desired final state within a selected maneuver time $t_{f}$, as computed using the methodologies detailed in \cite{boonrath2025mission}. The initial states for the minimum-energy optimal control problem correspond to the initial states of the \acp{MU}, while the final states are defined as follows. First, an aiming point is established for each \ac{MU}, on a plane that is located 5 m behind the debris's \ac{CoM} and is parallel to the $\hat{\mathbf{i}}$-$\hat{\mathbf{j}}$ plane (see Figure \ref{fig:tethernet_aim}). The $i$-th \ac{MU} aiming point defined as: 
\begin{equation}
\label{eq:aim}
\begin{aligned}
 \mathbf{r}_{\text{final},i}=(x_{\text{nom},i}+\Delta x_{i})\mathbf{\hat{i}}+(y_{\text{nom},i}+\Delta y_{i})\mathbf{\hat{j}}&+(z_{\text{D}}-5)\mathbf{\hat{k}}, \\& i=1,2,3,4 
\end{aligned}
\end{equation}
where $x_{\text{nom},i}$ and $ y_{\text{nom},i}$ are the nominal coordinates of the $i$-th \ac{MU} aiming point on the $\hat{\mathbf{i}}$-$\hat{\mathbf{j}}$ plane, which are defined to be $x_{\text{nom},i} = x_{\text{D}} + 12.0(-1)^i$ and $x_{\text{nom},i} = x_{\text{D}} + 12.0(-1)^{\lfloor (i+1)/2 \rfloor}$, while $x_{\text{D}}$, $y_{\text{D}}$, and $z_{\text{D}}$ are the debris's \ac{CoM} coordinates. The variables $\Delta x_i$ and $\Delta y_i$ denote coordinate offsets of the aiming point of the $i$-th \ac{MU}, to be optimized for by the \ac{GNN}-based optimization process proposed in this work. Next, the final velocity for each $i$-th \ac{MU}, $\mathbf{v}_{\text{final},i}$, is defined as: 
\begin{equation}
\label{eq:aimVel}
 \mathbf{v}_{\text{final},i} = \left(\frac{||\mathbf{r}_{\text{final},i}-\mathbf{r}_{0,i}||}{t_f}+v_i\right)\frac{\mathbf{r}_{\text{final},i}-\mathbf{r}_{0,i}}{||\mathbf{r}_{\text{final},i}-\mathbf{r}_{0,i}||}
\end{equation}
where $\frac{||\mathbf{r}_{\text{final},i}-\mathbf{r}_{0,i}||}{t_f}$ is the nominal final speed for the \acp{MU} and $v_i$ is a scalar quantity, selected via the \ac{GNN}-based optimization process, defining a discrepancy from the nominal final speed. For each $i$-th \ac{MU}, solving the optimal minimum-energy control problem with the initial and final states defined above results in the states $\mathbf{x}_{r, i}(t)=[\mathbf{r}_{{r,i}}^T(t) , \mathbf{v}_{{r,i}}^T(t) ]^T$ defined for the entire reference path, consisting of the reference positions $\mathbf{r}_{r, i}(t)$ and the reference velocities $\mathbf{v}_{r, i}(t)$.

To track the reference path, the control force for each $i$-th \ac{MU} is determined via a \ac{MPC} controller. To accomplish this, a reference \ac{MU} model without the net attached (i.e., the same model used to compute the minimum-energy reference path) is employed. This model and reference trajectory are converted from their original continuous-time form to a discrete-time form (with a sampling frequency of 20 Hz \cite{boonrath2024learning}) using the \texttt{python-control} library, yielding the discrete-time \(A\) and \(B\) matrices for the state and control, respectively. With the discrete-time model established, the mathematical formulation for the \ac{MPC} is defined as follows for each $i$-th \ac{MU}:
%To do so, a reference \ac{MU} model without the net attached (i.e., the same model that is used to compute the minimum-energy reference path) is employed as an approximate system model for \ac{MPC}, and is then converted into a discrete-time model through the use of the \texttt{python-control} library, resulting in discrete-time $A$ and $B$ system state and control matrices, respectively. Given the discrete-time model, the mathematical formulation for \ac{MPC} is described as follow:

\begin{equation}
\begin{aligned}
\min_{\{\mathbf{x}_{k,i},\mathbf{u}_{k,i}\}} \quad \label{eq:mpc}
& (\mathbf{x}_{N_h,i} - \mathbf{x}_{N_h,r,i})^T Q_N (\mathbf{x}_{N_h,i} - \mathbf{x}_{N_h,r,i}) 
\\ & \quad+ \sum_{k=0}^{N_h-1} \Big[ (\mathbf{x}_{k,i} - \mathbf{x}_{k,r,i})^T Q (\mathbf{x}_{k,i} - \mathbf{x}_{k,r,i}) \\
& \quad\quad + \mathbf{u}_{k,i}^T R \mathbf{u}_{k,i} \Big] \\
\text{subject to} \quad 
& \mathbf{x}_{k+1,i} = A \mathbf{x}_{k,i} + B \mathbf{u}_{k,i}, \quad k = 0, \dots, N_h-1 \\
% & \mathbf{x}_{\min,i} \leq \mathbf{x}_{k,i} \leq \mathbf{x}_{\max,i}, \quad k = 0, \dots, N \\
& \mathbf{u}_{\min,i} \leq \mathbf{u}_{k,i} \leq \mathbf{u}_{\max,i}, \quad k = 0, \dots, N_h-1 
%& \mathbf{x}_{0,i} = \bar{\mathbf{x}}_i
\end{aligned}
\end{equation}

\noindent
where the subscript $k$ indicates the $k$-th time step for \ac{MPC}\footnote{It should be noted that the actual net capture simulation is still performed in continuous time, whereas the above discrete-time formulation is only used for the MPC controller.}. States of each $i$-th \ac{MU} at each $k$-th step, $\mathbf{x}_{{i,k}} =[\mathbf{r}_{{i,k}}^T , \mathbf{v}_{{i,k}}^T ]^T $, are unconstrained for this work, while the inputs (i.e., the control force) $\mathbf{F}_{T,i,k} =\mathbf{u}_{i,k} = [u_{{x,i,k}}, u_{{y,i,k}}, u_{{z,i,k}}]^T$ are constrained as $\mathbf{u}_{\min,i} = -\frac{1}{\sqrt{3}}[F_{T,\text{max}}, F_{T,\text{max}}, F_{T,\text{max}}]^T$ and $\mathbf{u}_{\max,i} = \frac{1}{\sqrt{3}}[F_{T,\text{max}}, F_{T,\text{max}}, F_{T,\text{max}}]^T$ such that the total maximum thrust magnitude available for each \ac{MU} is $F_{T,\text{max}}$. The weight matrices are heuristically defined as $Q_N=Q =$diag([10.0, 10.0, 10.0, 1.0, 1.0, 1.0]) and $R=$diag([0.01, 0.01, 0.01]) with the finite-horizon step size, $N_h$, chosen for \ac{MPC} being 10. Using the \texttt{osqp} package in Python, Eq. \eqref{eq:mpc} is solved iteratively throughout the net's flight to the target at 20 Hz (i.e., the thrust value applied to the continuous-time, full-net system is updated every 0.05 s). With the \ac{MPC}-computed thrust values, the total fuel consumed is determined as:
\begin{equation}
\label{eq:RndFuel}
m_{\text{prop}} = \sum _{i = 1} ^{4} \int_0 ^{t_{f}} \frac{||\mathbf{F}_{T,i}(\tau) ||}{g_e I_{\text{sp}}} d\tau
\end{equation}
where $g_e=9.81$ m/s$^2$ and $I_{\text{sp}}$ is the specific impulse, whose value is dependent on the propulsion system to be chosen. 

\subsection{GNN-aided Optimization Formulation}
\begin{table*}[ht]
% \footnotesize
  \centering
  \caption{Tether-Net Fuel Consumption Optimization Combinatorial and Integer Variables}
  \label{tab:all_vars}
  \begin{minipage}{0.5\textwidth}
    \centering
    \caption*{Continuous Variables}
    \begin{threeparttable}
    \begin{tabular}{l|lllll}
      % \toprule
      \hline
      Category & Sub-category & Variable & Bound \\
      % \midrule
      \hline
      \multirow{3}{*}{Control}
      & \multirow{3}{*}{\ac{MU} Control}
      & $\Delta x_{i,i=1,2,3,4} $ [m] & [-5, 5] \\
      & &$\Delta y_{i,i=1,2,3,4} $ [m] & [-5, 5] \\
      & &$v_{i,i=1,2,3,4} $ [m/s] & [-1, 4]
      \\
      % \midrule
      \hdashline
      % \cdashline{2-4}[0.8pt/2pt]
      \multirow{5}{*}{Physical Design}
      & \multirow{1}{*}{\ac{MU} Mass}
      & $m_{\text{MU}} $ [kg] & [2, 3]
      \\
      % \hline
      % \hdashline
      \cdashline{2-4}[0.8pt/2pt]
      & \multirow{3}{*}{Net Morphology}
      & $r_{\text{thread}}$ [m] & [0.0005, 0.0015] \\
      & &$r_{\text{corner}}$ [m] & [0.0001, 0.0015] \\
      & &$L_{\text{net}}$ [m] & [19, 25]\\
      & &$L_{\text{ct}}$ [m] & [0.5, 2]
      \\
      % $N_k$ & Integer & \{9, 11, 13, 15\} & - \\
      % \bottomrule
      \hline
    \end{tabular}
    \begin{tablenotes}
      \item[] $\quad$ % dummy gap
    \end{tablenotes}
    \end{threeparttable}
    \label{tab:cont_vars}
  \end{minipage}
  % \hspace{0.5cm}
  % \vspace{3cm}
  % \begin{minipage}{0.8\textwidth}
  %   \centering
  %   \caption*{Combinatorial and Integer Variables}
  %   \begin{threeparttable}
  %   \begin{tabular}{l|llllllll}
  %     \hline
  %     Category & Sub-category & Variable & \#1 & \#2 & \#3 & \#4 & \#5 \\
  %     % \midrule
  %     \hline
  %     \multirow{5}{*}{Combinatorial}
  %     & \multirow{3}{*}{Thruster} 
  %     & $F_{T,\text{max}}$ [N] & 8.9 & 3.6 & 6.1 & 5.5 & 6.0 \\
  %     & & $I_{\text{sp}}$ [s] & 60.0 & 57.0 & 277.0 & 253.0 & 250.0 \\
  %     & & $m_T$ [kg] & 0.37 & 0.023 & 0.6 & 0.48 & 0.25 \\ 
  %     % \hdashline
  %     \cdashline{2-8}[0.8pt/2pt]
  %     & \multirow{2}{*}{Net Material}
  %     & $E_{\text{n}}$ [Gpa] & 70.0 & 70.5 & 112.4 & - & -\\
  %     & & $\rho_{\text{n}}$ [kg/m$^3$] & 1390.0 & 1440.0 & 1440.0 & - & -\\ 
  %     \hdashline
  %     \multirow{2}{*}{Integer}
  %     & \multirow{2}{*}{Net Mesh}\\
  %     & $N_k$ for square & 9 & 11 & 13 & - & -\\
  %     & $N_k$ for hex & 7 & 9 & 11 & - & -\\
  %     % \hdashline
  %     \cdashline{2-8}[0.8pt/2pt]
  %     & \multirow{1}{*}{Closing Node Index}
  %     & $K_{\text{cls}}$ & 0-3 & 0-4 & 0-5 & - & -\\
  %     % \bottomrule
  %     \hline
  %   \end{tabular}
  %   \end{threeparttable}
  %   \label{tab:comb_vars}
  % \end{minipage}
  \begin{minipage}{0.8\textwidth}
    \centering
    \caption*{Combinatorial and Integer Variables (From \cite{gnn_reco_idetc})}
    \begin{threeparttable}
    \begin{tabular}{l|lllllllll}
      \hline
       Category & Sub-category & Variable & \#1 & \#2 & \#3 & \#4 & \#5 \\
      % \midrule
      \hline
      \multirow{8}{*}{Combinatorial}
      & \multirow{3}{*}{Thruster} 
      & $F_{T,\text{max}}$ [N] & 8.9 & 3.6 & 6.1 & 5.5 & 6.0 \\
      & & $I_{\text{sp}}$ [s] & 60.0 & 57.0 & 277.0 & 253.0 & 250.0 \\
      & & $m_T$ [kg] & 0.37 & 0.023 & 0.6 & 0.48 & 0.25 \\ 
      % \hdashline
      \cdashline{2-9}[0.8pt/2pt]
      & \multirow{2}{*}{Net Material}
      & $E_{\text{n}}$ [GPa] & 70.0 & 70.5 & 112.4 & - & - \\
      & & $\rho_{\text{n}}$ [kg/m$^3$] & 1390.0 & 1440.0 & 1440.0 & - & - \\ 
      \cdashline{2-9}[0.8pt/2pt]
      & \multirow{2}{*}{Net Shape}
      % & $N_{\text{MU}}$ & 4 & 4 & 4 & 6 & 6 & 6\\
      & $N_k$ & 9 & 11 & 13 & - & - \\
      & & $K_{\text{cls}}$ & -2 to 0 & -2 to +1 & -3 to +1 & - & - \\
      % \hdashline
      % \multirow{2}{*}{Integer}
      % & \multirow{2}{*}{Net Mesh}\\
      % & $N_k$ for square & 9 & 11 & 13 & - & -\\
      % & $N_k$ for hex & 7 & 9 & 11 & - & -\\
      % % \hdashline
      % \cdashline{2-8}[0.8pt/2pt]
      % & \multirow{1}{*}{Closing Node Index}
      % & $K_{\text{cls}}$ & 0-3 & 0-4 & 0-5 & - & -\\
      % \bottomrule
      \hline
    \end{tabular}
    \end{threeparttable}
    \label{tab:comb_vars}
  \end{minipage}
  
\end{table*}
For tether-net systems, the physical design of the net and the controlled trajectory of the \acp{MU} significantly affect the probability of capture success. However, most previous works \cite{jang2023active,jsr2023Zeng} that examined design and control variables used continuous or discrete integer values, with limited studies exploring physical designs involving combinatorial variables. Considering combinatorial variables is equally important, as certain design choices inherently involve fixed parameter groupings rather than independent continuous values. 

The goal of the optimization task is to minimize the fuel cost, in terms of propellant mass consumed $m_{\text{prop}}$, during the capture mission while accomplishing a successful capture. The fuel cost is calculated based on the thrust of the \acp{MU}, which is controlled by the \ac{MPC} controller, and the maximum thrust is specified by the manufacturer. All the related optimization variables are listed in Table \ref{tab:all_vars}. 

\textbf{Continuous Variables:} Continuous variables include the \ac{MU} control variables, the mass of each \ac{MU}, and the net morphology parameters. The \ac{MU} control variables are defined by the aiming point position illustrated in Fig. \ref{fig:tethernet_aim} and the final speed of the \ac{MU} (i.e., the speed at $t_f$). The net morphology parameters include the net thread radius $r_{\text{net}}$, the corner thread radius $r_{\text{corner}}$, the length of net thread $L_{\text{net}}$, and the length of the corner thread $L_{\text{ct}}$. These parameters are depicted in the top-left part of Fig. \ref{fig:close_node}. 

\textbf{Combinatorial and Integer Variables: }
Five thrusters were picked as the potential choices, and each thruster is defined by its maximum thrust magnitude $F_{T,\text{max}}$, its specific impulse $I_{\text{sp}}$, and its mass $m_T$; the considered thrusters include cold-gas and combustion-based thrusters in order to explore the design space of the tether-net system \cite{nasa2025soa, ecaps_brochure_2023, rubicon_5n_lt_thruster}. Three possible types of high-strength fibers are picked for the threads of the net \cite{BottaPhdThesis,kevlar_technical_guide}, each defined by the material's Young's Modulus $E_{\text{n}}$, and density $\rho_{\text{n}}$; the Poisson ratio for all fibers is assumed to be the same. 
The variable $N_k$ is an integer variable that indicates the number of knots on each side of the net, resulting in a net with $N_k^2$ knots in total. The variable $K_{\text{cls}}$ represents the index of the node on a net edge where the closing mechanism is connected, as shown on the right side of Fig. \ref{fig:close_node}. A higher index value corresponds to a connection node closer to the central node of the edge\footnote{The set of connection nodes selected by default (i.e., the connection nodes selected when $K_{\text{cls}}=0$) is chosen according to the methodology presented in \cite{botta2019simulation}.}. Note that 1) the upper and lower bounds of $K_{\text{cls}}$ depend on $N_k$ (see Table \ref{tab:comb_vars}), and 2) the same $K_{\text{cls}}$ value results in different offsets from each edge's central node when $N_k$ differs. For all $K_{\text{cls}}$, the connection node offsets are defined so the two closing nodes of each net edge are displaced equally away from the center of each edge (i.e., they are symmetric). 
%The variable $K_{\text{cls}}$ represents the index of the node on an edge of the net where the closing mechanism is connected\footnote{The set of connection nodes selected by default (i.e., the connection nodes selected when $K_{\text{cls}}=0$) is chosen according to the methodology presented in \cite{botta2019simulation}.}, as shown on the right side of Fig. \ref{fig:close_node}; a higher index value corresponds to a connection node located that is located closer to central node of the edge. 
%It should be noted that 1) the upper and lower bounds of $K_{\text{cls}}$ depend on $N_k$ (see Table \ref{tab:comb_vars}) and that 2) the same value of $K_{\text{cls}}$ results in different offsets from the central node of the edge when $N_k$ differs. For all $K_{\text{cls}}$, the connection node offsets are defined so that the two closing nodes of each net edge are displaced by the same amount away from the center of each edge (i.e., they are symmetric). 
%It should be noted that the upper bound of the value of $K_{\text{cls}}$ depends on $N_k$, as Table \ref{tab:comb_vars} shows. Additionally, as the net mesh number $N_k$ varies, the distance between the closing nodes changes; hence, the same $K_{\text{cls}}$ value implies different physical representations when $N_k$ differs. Thus, $K_{\text{cls}}$ implies $(4+5+6)$ unique choices of the closing node position. 
In total, 180 unique combinations exist for thruster, net material, and net morphology variables. Based on previous research \cite{gnn_reco_idetc}, for a problem with this many candidate combinations, the use of \ac{GNN} improves optimization efficiency by quickly identifying the best-suited combinatorial variables. The optimization is then defined as the following:

% \begin{table*}[ht!]
% \footnotesize
% \centering
% \begin{threeparttable}
% \caption{Bounds of the Random Variables for Sampling}
% \label{tab:bound}
% \begin{tabular}{llllll}
% \toprule
% & Variables & Definition &Unit & Lower Bound & Upper Bound \\
% \midrule
% \multirow{3}{*}{Target States} 
% & $X_D, Y_D, Z_D$ & Position coordinates & m & -9, -9, -60 & 9, 9, -40 \\
% & $O_{D_{x0}}, O_ {D_{y0}}, O_{D_{z0}}$ & 3-2-1 Euler angles & deg & -180, -180, -180 & 180, 180, 180 \\
% & $\omega_{D_{x0}}, \omega_{D_{y0}}, \omega_{D_{z0}}$ & Angular velocity components & deg/s & 1, 1, 5 & 10, 10, 30 \\
% % \midrule
% \bottomrule 
% \end{tabular}
% \end{threeparttable}
% % \vspace{-15pt}
% \end{table*}

% To obtain a more generalized, optimized design, we propose sampling 10 scenarios, defined by the target states in Table \ref{tab:bound}, using \ac{LHS}. These 10 scenarios will be used to generate the dataset for training the \ac{GNN}, which will be introduced in Section \ref{sec:results}, and will also be used for the optimization function evaluations. 

\begin{equation}
\footnotesize
\label{eq:obj_raw}
\begin{aligned}
 \min_{\mathbf{Z}_{\text{comb}}, \mathbf{X}_{\text{cont}}} &f( \mathbf{Z}_{\text{comb}}, \mathbf{X}_{\text{cont}} ) = f_{\text{MU}} \\
\text{s.t.} \quad & g(\mathbf{Z}_{\text{comb}}, \mathbf{X}_{\text{cont}}) = I^*_{\text{CQI}} \leq 2.5\\
& h(\mathbf{Z}_{\text{comb}}, \mathbf{X}_{\text{cont}})= N_{L} = 12\\
\text{where} \quad 
& f_{\text{MU}} = m_{\text{prop}}\\
& \mathbf{Z}_{\text{comb}} = \begin{bmatrix} 
F_{T,\text{max}}, I_{\text{sp}}, m_T, E_{\text{n}}, \rho_{\text{n}}, N_k, K_{\text{cls}}\end{bmatrix} \\
& \mathbf{X}_{\text{cont}} =
\begin{bmatrix}
\begin{aligned}
& \Delta x_{1} \dots \Delta x_{4},\ \Delta y_{1} \dots \Delta y_{4}, \\
& v_{1} \dots v_{4},\ m_{\text{MU}},\ r_{\text{thread}},\ r_{\text{corner}},\ L_{\text{net}},\ L_{\text{ct}}
\end{aligned}
\end{bmatrix}\\
% & x_n^L \leq x_n \leq x_n^U, \quad \forall ~n=1, 2, \dots, k_{\text{cont}}\\
& \mathbf{Z}_{\text{comb}} \in \mathbb{Z}
\end{aligned}
\end{equation}
$\mathbb{Z} = [\mathbf{Z}^{(1)}, \mathbf{Z}^{(2)}, \dots, \mathbf{Z}^{(180)}]$ is the set of all the 180 combinations of the combinatorial and integer variables in Table \ref{tab:comb_vars}. However, when dealing with a mixed-combinatorial problem with this many combinations, it would be too time-consuming to run the function evaluations on each combination, especially when the objective and constraints can only be calculated after completion of the time-expensive, relatively high-fidelity simulation of the system dynamics. Therefore, it is beneficial to have an approximator that can quickly approximate which combination is the best-suited one for the given vector of continuous variables, so that the combinatorial and integer variables no longer need to be co-optimized with the continuous variables. Assuming the approximator is represented by $\Theta$, the objective function can be transformed into the following: 

\begin{equation}
\footnotesize
\label{eq:obj_mod}
\begin{aligned}
 \min_{\mathbf{X}_{\text{cont}}} &f( \mathbf{Z}^*_{\text{i}}, \mathbf{X}_{\text{cont}} ) = f_{\text{MU}} \\
\text{s.t.} \quad & g(\mathbf{Z}, \mathbf{X}_{\text{cont}}) = I^*_{\text{CQI}} \leq 2.5\\
& h(\mathbf{Z}, \mathbf{X}_{\text{cont}})= N_{L} = 12\\
\text{where} \quad 
& f_{\text{MU}} \left(\mathbf{Z}_{\text{comb}}, \mathbf{X}_{\text{cont}} \right)= m_{\text{prop}} \\
& \mathbf{y} = {\Theta}(\mathcal{G}, \mathbf{X}_{\text{cont}}) \\
& i^* = \underset{i=1, 2, \dots, n_{\text{tot}}}{\text{argmax}} y_i, \quad y_i \in \mathbf{y}\\
& \mathbf{X}_{\text{cont}} =
\begin{bmatrix}
\begin{aligned}
& \Delta x_{1} \dots \Delta x_{4},\ \Delta y_{1} \dots \Delta y_{4}, \\
& v_{1} \dots v_{4},\ m_{\text{MU}},\ r_{\text{thread}},\ r_{\text{corner}},\ L_{\text{net}},\ L_{\text{ct}}
\end{aligned}
\end{bmatrix}\\
% & x_n^L \leq x_n \leq x_n^U, \quad \forall ~n=1, 2, \dots, k_{\text{cont}}\\
& \mathbf{Z}^*_{i} \in \mathcal{G},~{\text{and}} ~\mathcal{G}\in\mathbb{Z}
\end{aligned}
\end{equation}
Here, $\mathbf{y}$ is a vector of scores of each node in the sampled subgraph, and $\mathcal{G}$ is a \textit{graph} constructed by all the combinations, which will be discussed more in the following section. To handle the constraints, we propose using a penalty function to simplify the objective function, which can then be further transformed into the following:
% \begin{equation}
% \label{eq:obj_MU}
% \begin{aligned}
% \min_{\mathbf{X}_{\text{cont}}} &f( \mathbf{Z}^*_{\text{comb}}, \mathbf{X}_{\text{cont}} ) = f_{\text{MU}} \\
% \text{where} \quad & \mathbf{Z}^*_{\text{comb}} = 
% \mathcal{Z}_{\text{argmax}(\mathbf{y})} \\
% & \mathbf{y} = \text{GNN-ReCo}(\mathcal{G}, \mathbf{X}_{\text{cont}}) \\
% & f_{\text{MU}} = \begin{cases} 
%     \sum_{j=1}^{10}m_{\text{prop},j} \quad,\quad \text{if } I_{\text{CQI},j}^* < 2.5 \text{ and } N_{L,j} = 12, \forall~j=1,2,...,10 &\\
%     \sum_{j=1}^{10} \ln\left( (I_{\text{CQI},j}^*-2.5)^2+1\right) +
%     \ln \left((N_{\text{L},j}-12)^2+1\right) + \left(\gamma_{\text{tension},j}-1\right) + \beta \quad,\quad \text{otherwise} \\ \quad\quad\quad\quad\quad\quad\quad\quad\quad\quad\quad\quad\quad\quad\quad\quad\quad\quad\quad\quad\quad\quad\quad\quad\quad\quad\quad\quad\forall~j=1,2,...,10&
%   \end{cases} \\
% & \mathbf{X}_{\text{cont}} =\begin{bmatrix} \Delta x_{1} \dots \Delta x_{4} & \Delta y_{1} \dots \Delta y_{4} & v_{1} \dots v_{4} & m_{\text{MU}} & r_{\text{thread}} & r_{\text{corner}} & L_{\text{net}} & L_{\text{ct}}
% \end{bmatrix}
% \end{aligned}
% \end{equation}
\begin{equation}
\footnotesize
\label{eq:obj_MU}
\begin{aligned}
 \min_{\mathbf{X}_{\text{cont}}} &f( \mathbf{Z}^*_{\text{comb}}, \mathbf{X}_{\text{cont}} ) = f_{\text{MU}} \\
\text{where} \quad & \mathbf{Z}^*_{\text{comb}} = 
\mathcal{Z}_{\text{argmax}(\mathbf{y})} \\
& \mathbf{y} = \text{GNN-NavCo}(\mathcal{G}, \mathbf{X}_{\text{cont}}) \\
& f_{\text{MU}} \left(\mathbf{Z}_{\text{comb}}, \mathbf{X}_{\text{cont}} \right) = \begin{cases} 
     m_{\text{prop}},~\text{if } I_{\text{CQI}}^* \leq 2.5 \text{ and } N_{L} = 12 &\\
    \ln\left( (I_{\text{CQI}}^*-2.5)^2+1\right) &\\ \quad\quad +
    \ln \left((N_{\text{L}}-12)^2+1\right) + \beta, \text{otherwise} \\
  \end{cases} \\
& \mathbf{X}_{\text{cont}} =
\begin{bmatrix}
\begin{aligned}
& \Delta x_{1} \dots \Delta x_{4},\ \Delta y_{1} \dots \Delta y_{4}, \\
& v_{1} \dots v_{4},\ m_{\text{MU}},\ r_{\text{thread}},\ r_{\text{corner}},\ L_{\text{net}},\ L_{\text{ct}}
\end{aligned}
\end{bmatrix}\\
\end{aligned}
\end{equation}
\noindent where \ac{GNN}-NavCo is the \ac{GNN} recommender that will be introduced in the following section, $\mathbf{y}$ is a vector of scores of each node in the sampled subgraph, and $\beta$ is a penalty term, equal to 1.5 times the maximum fuel cost observed in a successful capture\footnote{A failed capture was found to generally have a worse value of $f_{\text{MU}}$ than a successful capture.}. The simulator includes a maximum tension tolerance that checks whether the tension in the net thread exceeds the upper limit; readers may refer to \cite{boonrath2025robustness} for details on how the tension limit is defined. If tension in any thread exceeds the maximum, then the simulation is terminated and tagged as unsuccessful capture, and the values of $I_{\text{CQI}}^*$ and $N_{L}$ are set to be 50.0 and 0, respectively, which are the worst observed values during preliminary simulations.

\section{Graph-Learning-Based Recommender for the Optimizer}\label{sec:opt}

\subsection{Preprocessing and Creation of the Graph Dataset}
The goal of the \ac{GNN} is to act as a recommender, which, given the input of a graph of nodes and a continuous context, can rank them and give the best combinatorial node for that context. In this research, the \ac{GNN} takes an undirected graph and a vector of continuous variables as input and outputs a directed graph whose edges encode the pairwise differences between nodes with respect to the objective function used to train the \ac{GNN}. In this concept, graph $\mathcal{G}$ is defined to be $\mathcal{G} = \{\mathcal{Z}, \mathcal{E}\}$, where $\mathcal{Z}$ is the node feature and $\mathcal{E}$ is the edge feature. In this work, assuming there are $k_{\text{comb}}$ combinatorial variables, then a unique node is defined to be $ \mathbf{Z} = [ z_1, z_2, \dots, z_{k_{\text{comb}}} \mid z_j \in \mathbb{R}, \, \, j = 1, 2, \dots, k_{\text{comb}} ]$, which represents the direct features of a combination in the $k_{\text{comb}}$ dimensional space, and $z_{j}$ is the $j$-th combinatorial variable. Thus, the node feature $\mathcal{Z} \in \mathbb{R}^{n_{\text{tot}} \times k_{\text{comb}}}$ is a vector containing all predefined valid combinations of the combinatorial variables. All the nodes are defined to be connected with each other in this graph to form a fully connected graph \cite{GNN_distill}, and the edge feature $\mathcal{E} \in \mathbb{R}^{n_{\text{tot}} \times n_{\text{tot}}}$ is then defined as an adjacency matrix between each node encoding the pairwise differences of the objective function between them, parametrized by the continuous context.

For a complex problem, the combinatorial variable number $k_{\text{comb}}$ and the number of valid combinations $m$ can be large. In that case, the dimensions of the node feature $\mathcal{Z}$ and edge feature $\mathcal{E}$ are large for training the \ac{GNN} and for further analysis. Therefore, instead of directly working on the graph $\mathcal{G}$ with all the nodes and edges, we propose to randomly pick $n_{\text{sn}}$ nodes from $\mathcal{Z}$ and their corresponding edges from $\mathcal{E}$ to create a set of $p_{\text{sg}}$ subgraphs. The set of subgraphs $S$ is defined as:
\begin{equation}
  S = \{\mathcal{G}_1, \mathcal{G}_2, \dots, \mathcal{G}_{p_{\text{sg}}} \mid \mathcal{G}_q \subset \mathcal{G}, \, \forall ~q = 1, 2, \dots, p_{\text{sg}} \}
\end{equation}
where $p_{\text{sg}}$ is the number of subgraphs, and each subgraph $\mathcal{G}_q$ in the set is defined to be:
\begin{align}
  \mathcal{G}_q &= \{\mathcal{Z}_q, \mathcal{E}_q \mid \mathcal{Z}_q \in \mathcal{Z}, \mathcal{E}_q \in \mathcal{E}, \, \forall ~q = 1, 2, \dots, p_{\text{sg}} \} \notag \\
  & \mathcal{Z}_q = \begin{pmatrix}
\mathbf{Z}_{q,1} \\
\mathbf{Z}_{q,2} \\
\vdots \\
\mathbf{Z}_{q, n_{\text{sn}}}
\end{pmatrix} , \, \, 
\mathcal{E}_q = \begin{pmatrix}
\mathbf{d}_{q,1} \\
\mathbf{d}_{q,2} \\
\vdots \\
\mathbf{d}_{q, n_{\text{sn}}}
\end{pmatrix}, 
\\
&\quad \mathbf{Z}_{q,i} \in \mathcal{Z}, \mathbf{d}_{q, i} \in \mathcal{E}, \, \forall ~i = 1, 2, \dots, n_{\text{sn}} \notag
\end{align}
where the notation $(\cdot)_{q,i}$ denotes the operation of randomly picking an unrepeated value between $[1, n_{\text{tot}}]$. The variable $\mathbf{d}_i$ is the vector of pairwise differences in objective value parametrized by the continuous context of the $i$-th node to every node in the graph.

Then, within a subgraph $\mathcal{G}_q$, the objective function will be evaluated $n_{\text{sn}}$ times, with each node providing the combinatorial variables, and $\mathbf{X}_{q}$ providing the continuous variables; thus, all nodes use the same vector of continuous values for function evaluations. The objective values for all the nodes in $\mathcal{G}_q$ and the continuous variables in $\mathbf{X}_q$ is then defined as:
\begin{align}
  \mathbf{F}_q &= \begin{pmatrix}
f_{q,1} \\
f_{q,2} \\
\vdots \\
f_{q, n_{\text{sn}}}
\end{pmatrix}, \, f_{q,i} = f(\mathbf{Z}_{q,i}, \mathbf{X}_q) \\
&\quad \forall ~i = 1, 2, \dots, n_{\text{sn}} \notag
\end{align}

% GM: Review
To train the \ac{GNN} as a recommender (referred to as \ac{GNN}-NavCo in the rest of the paper) to find the best-suited combination, a pairwise regression loss function such as \ac{MSE} or Huber loss is used during training. Huber loss is a special loss function that acts as \ac{MSE} loss function below a threshold and \ac{MAE} loss above it. Here, the special cyclical consistency regularizer is also used, which penalizes nonzero circulation around graph cycles, encouraging globally consistent predictions. The variable $\lambda_{\text{cycle}}$ is the regularization coefficient that weights the cyclical consistency term relative to the primary pairwise regression objective. 
Let $\hat{f}(i,j;{\mathbf{X}})$ denote the predicted edge difference. Learning proceeds by minimizing the discrepancy between predicted and observed differences over the edges of the graph, where $\ell(\cdot)$ denotes a regression loss where edge-wise loss is defined in Eq.~\eqref{eq:edge_loss}. Mathematically, cycle consistency penalizes nonzero circulation over triangles, which are the smallest cycles in a fully connected graph. For a triangle $(i,j,k)$, the circulation is defined in Eq.~\eqref{eq:cycle_defintion}, and ideally $c(i,j,k;{\mathbf{X}}) = 0$. If $\mathcal{T}$ denotes the set of all triangles in the graph, then the cycle consistency loss is defined in Eq. \eqref{eq:cycle_regularizer}. Thus, the overall objective is $ L = L_{\text{edge}} + \lambda_{\text{cycle}} L_{\text{cycle}}$.

\begin{align}
\label{eq:edge_loss}
    L_{\text{edge}} = \mathbb{E}_{\mathbf{X}} \left[
\frac{1}{|\mathcal{E}_q(\mathcal{G}_q)|}
\sum_{(i,j)\in \mathcal{E}_q(\mathcal{G}_q)}
\ell\big(\hat{f}_{q}(i,j;{\mathbf{X}}) - f_{q}(i,j)\big)
\right] 
\end{align}

\begin{align}
\label{eq:cycle_defintion}
    c(i,j,k;{\mathbf{X}}) = \hat{f}(i,j;{\mathbf{X}}) + \hat{f}(j,k;{\mathbf{X}}) + \hat{f}(k,i;{\mathbf{X}}) 
\end{align}

\begin{align}
\label{eq:cycle_regularizer}
    L_{\text{cycle}} = \mathbb{E}_{\mathbf{X}} \left[
\frac{1}{|\mathcal{T}|}
\sum_{(i,j,k)\in \mathcal{T}} c(i,j,k;{\mathbf{X}})^2
\right] 
\end{align}

In terms of the tether-net optimization problem, the continuous variables in the top part of Table \ref{tab:all_vars} are used to define $\mathbf{X}_q$; the combinatorial and integer variables in the bottom part of Table \ref{tab:all_vars} are used to create the combinations, which then formulated into $\mathbf{Z}_{q}$.

\begin{figure*}[ht!]
  \centering
  \includegraphics[width=0.8\linewidth]{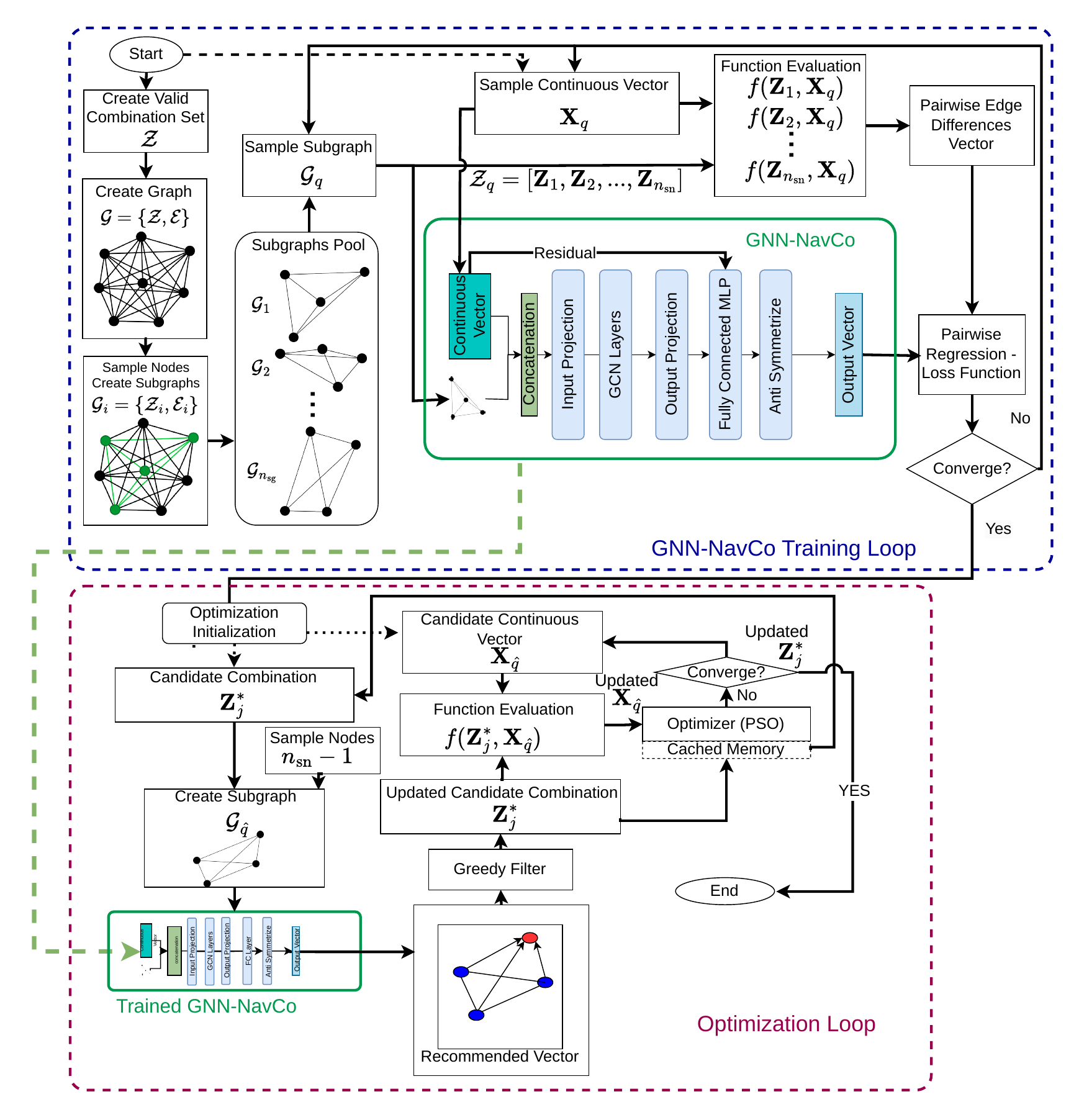}
  \caption{GNN-aided Optimization Framework}
  \label{fig:overall}
\end{figure*}

\begin{figure*}[ht!]
  \centering
  \includegraphics[width=0.9\linewidth]{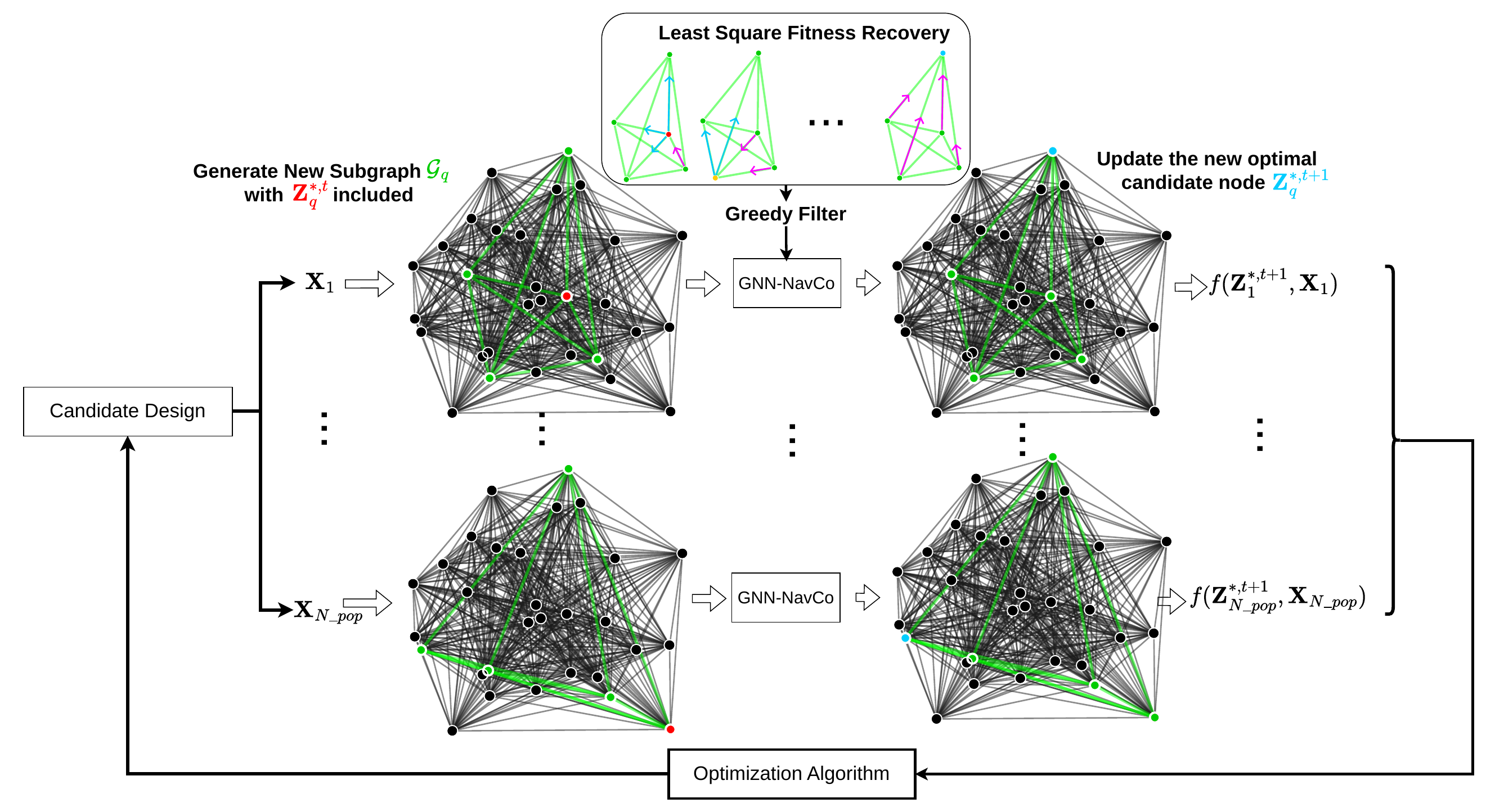}
  \caption{Sketch of the \ac{GNN}-NavCo-aided optimization using population-based optimizer as an example. The graphs are simplified for illustrative purposes. At the beginning of each iteration, a subgraph is sampled with the optimal node from the previous iteration (red node) included. During the Least Square Fitness Recovery process, the fitness score of each node (red and orange nodes) in the subgraph is evaluated based on the edge values. A greedy filter is applied to find the node with the optimal score (blue node). In the top figure, between two nodes in the subgraph, the blue arrow indicates the direction that leads to a node with a better score, and the purple arrow indicates the direction that leads to a node with a worse score. The optimal node, along with the optimized continuous vector $\textbf{X}$, will be used in the function evaluation in the optimization.}
  \label{fig:mdpso}
\end{figure*}

%Figure \ref{fig:overall} shows the overall structure of the proposed framework. The framework consists of two major components: the \ac{GNN}-NavCo training and the optimization loop. The framework starts with creating the dataset by randomly sampling valid combinations to create a pool of subgraphs for training the \ac{GNN}-NavCo. 

% To train \ac{GNN}-NavCo, a continuous vector is sampled, and the sampled subgraphs will provide $n_{\text{sn}}$ subgraph nodes, and the combinations in the subgraph are sent to acquire $n_{\text{sn}}$ ground truth fitness values. The sample graph and the continuous vector are also sent to \ac{GNN}-ReCo to get $n_{\text{sn}}$ prediction score. Then the scores and the fitness values are sent to the loss function to calculate the gradient and update \ac{GNN}-ReCo.

%% GM 

\subsection{Framework of GNN-aided Optimization Algorithm}\label{sec:frame}
% \begin{figure}[ht!]
%   \centering
%   \includegraphics[width=1.0\linewidth]{Images/gnn_navco_framework.pdf}
%   \caption{GNN-aided Optimization Framework}
%   \label{fig:overall}
% \end{figure}

Figure \ref{fig:overall} shows the overall structure of the proposed framework. The framework consists of two major components: the \ac{GNN}-NavCo training and the optimization loop. The framework starts with creating the dataset by randomly sampling valid combinations to create a pool of subgraphs for training the \ac{GNN}-NavCo. To train \ac{GNN}-NavCo, the input consists of the sampled subgraph providing $n_{\text{sn}}$ combinatorial nodes and a common continuous vector. After this, the fitness scores of the combination are computed and stored on the nodes. The subgraph is a complete graph, meaning all nodes are connected to each other. Here, we capture the pairwise difference in fitness scores at nodes and move it into the edges, forming the ground truth of the abstraction. Then the fitness scores, represented by the objective value $f_{\text{MU}}$ here, are sent to the loss function to calculate the gradient and update \ac{GNN}-NavCo. The main objective loss is a pairwise regression loss. Therefore, the \ac{GNN}-NavCo is trained to predict the edge-wise pairwise difference in fitness scores between nodes in a subgraph given a continuous vector; in other words, the \ac{GNN}-NavCo learns a context-conditioned parametric edge function over the subgraph that models pairwise objective differences between combinatorial choices. This gives rise to an inference time phenomenon, where we input the set of undirected subgraphs, with nodes containing the combinatorial vector and a continuous context, and \ac{GNN}-NavCo outputs a directed graph, with pairwise differences in fitness scores encoded on the edges. Using this, ideally, the optimizer can make an informed decision to move to the best combinatorial node within the subgraph for the fixed best continuous vector provided by an outer optimizer, such as PSO. Thus, it achieves a notion of explicit gradient awareness over combinatorial spaces.

After \ac{GNN}-NavCo is trained, it helps the optimizer select the candidate combination from a subgraph generated from sampled combinations and the candidate combination from the previous iteration. The sampled combinations are randomly sampled from the graph with respect to the candidate combination. Once the directed graph is encoded with pairwise fitness score differences, the node-level fitness scores are recovered by solving a least-squares problem that minimizes the discrepancy between the computed edge differences and the induced node fitness scores. This yields an approximate global ranking over nodes \cite{leastsquares}. After this, a greedy filter is applied to select the node with the lowest score. The selected candidate combination, $\textbf{Z}^*_{q}$, is used in the function evaluation with the continuous vector provided by the optimizer, and it is also stored in the cache, so it becomes the new initial candidate combination in the next iteration. The continuous vector will be updated by the optimizer. The optimization iterates until the maximum number of iterations is reached or the result converges. Here, the \ac{MDPSO} algorithm~\cite{chowdhury2013mixed, tong2016multi} is used as the optimizer, which is a popular approach for solving complex nonlinear optimization problems. However, it should be noted that the optimizer can be any gradient- or sampling-based algorithm, such as a Genetic Algorithm or sequential quadratic programming. To validate that the converged designs are local optima, after the optimization converges, the combinatorial variables are frozen, and a sequential quadratic programming solver is applied to the continuous variables for post-optimization. The trained \ac{GNN} can efficiently pick the node to run the function evaluations without iterating over all the candidates. This \ac{GNN}-NavCo-aided MDPSO is defined as in Algorithm \ref{algo:GNN_mdpso}, and the illustrative flowchart is provided in Figure \ref{fig:mdpso}. 

In this research, the population of the \ac{MDPSO} is set to 100, and the max iteration limit is set to 200. The optimization is terminated early if either of the following conditions is met: i) the population remains entirely infeasible and the total constraint violation of the global best does not decrease (within a tolerance of $10^{-5}$) over 15 consecutive iterations; or ii) the global best is feasible but its objective value does not improve (within a tolerance of $10^{-5}$) over 15 consecutive iterations. All the other hyperparameters of the \ac{MDPSO} are set to their default values.

\begin{algorithm}
\caption{GNN-aided MDPSO}
\label{algo:GNN_mdpso}
% \footnotesize
\begin{algorithmic}[1] % [1] means numbering of lines
\STATE Initialize the particle population $\mathbf{X}_1$ to $\mathbf{X}_{N_{\text{pop}}}$
\STATE Initialize the optimal node $\mathbf{Z}^*_1$ to $\mathbf{Z}^*_{N_{\text{pop}}}$
\STATE Function evaluation $f(\mathbf{Z}^*_i, \mathbf{X}_i)$, $i= 1$ to $N_{\text{pop}}$
\FOR{$t = 1$ to $max\_iteration$}
  \STATE Update particles $\mathbf{X}_1$ to $\mathbf{X}_{N_{\text{pop}}}$ with \ac{MDPSO}
  \FOR{$q=1$ to $N_{\text{pop}}$}
    \STATE Select $n_{\text{sn}}-1$ nodes randomly around each $\mathbf{Z}^*_q$ from $\mathcal{G}$
    \STATE Create $\mathcal{G}_q$ 
    \STATE run $\mathbf{\mathcal{G}}^*_q= $ \ac{GNN}-NavCo($\mathcal{G}_q, \mathbf{X}_q$)
     \STATE \# \textit{recover node score from adjacency matrix } 
    \STATE $\Phi_i = \frac{1}{n_{\text{sn}}} \sum_{j=1}^{n_{\text{sn}}} \mathcal{E}^*_{ij}$
    
    \STATE Set $\textbf{Z}^*_q = \arg\min_i \Phi_i $ 
  \ENDFOR
  \STATE Function evaluation $f(\mathbf{Z}^*_i, \mathbf{X}_i)$, $i= 1$ to $N_{\text{pop}}$

  \IF{Converge}
    \STATE Return the result
  \ENDIF
\ENDFOR
\STATE Return the result
\end{algorithmic}
\end{algorithm}

% \begin{table*}[ht!]
% % \footnotesize
% \centering
% \begin{threeparttable}
% \caption{Scenario Information - Target State}
% \label{tab:init_state}
% \begin{tabular}{lllll}
% \toprule
% & Variables & Definition &Unit & Value \\
% \midrule
% % \multirow{3}{*}{Target States} 
% & $X_D, Y_D, Z_D$ & Position coordinates & m & 9, 9, -60 \\
% & $O_{D_{x0}}, O_ {D_{y0}}, O_{D_{z0}}$ & 3-2-1 Euler angle initial orientations & deg & 60, 40, 0 \\
% & $\omega_{D_{x0}}, \omega_{D_{y0}}, \omega_{D_{z0}}$ & Angular velocity components & deg/s & 10, 10, 30 \\
% % \midrule
% % \hdashline
% % \multirow{3}{*}{\ac{MU} controls} 
% % & $\psi_{T,1}, \psi_{T,2}, \psi_{T,3}, \psi_{T,4}$ & $\mathbf{\hat{i}} - \mathbf{\hat{j}}$ thrust angles for $i$-th \ac{MU} & deg & -45, -45, -45, -45 & 135, 135, 135, 135 \\
% % & $\theta_{T,1}, \theta_{T,2}, \theta_{T,3}, \theta_{T,4}$ & $\mathbf{\hat{k}} - \mathbf{\hat{j}}$ thrust angles for $i$-th \ac{MU} & deg & 35, 35, 35, 35 & 55, 55, 55, 55 \\
% % & $F_{T,\text{max}}$ & Thrust magnitude & N & 5 & 12 \\
% \bottomrule 
% \end{tabular}
% \end{threeparttable}
% \end{table*}

\subsection{MDPSO without the aid of the GNN-NavCo}
To investigate the contribution of \ac{GNN}-NavCo to the optimization, another round of \ac{MDPSO} is performed without \ac{GNN}-NavCo and uses an integer variable ($x_{\text{int}} \in [1, 180]$) to represent the choice of the combination among the feasible set. All the settings of this \ac{MDPSO} are exactly the same as the GNN-NavCo-aided \ac{MDPSO}. The initial populations, including both the continuous and projected combinatorial variables, are also kept the same between the two methods. The objective function for this round of optimization is defined in Eq. \eqref {eq:obj_MU_nognn}. The convergence history and optimized results are provided in the following Results section.

\begin{equation}
\footnotesize
\label{eq:obj_MU_nognn}
\begin{aligned}
 \min_{x_{\text{int}}, \mathbf{X}_{\text{cont}}} &f( x_{\text{int}}, \mathbf{X}_{\text{cont}} ) = f_{\text{MU}} \\
\text{where} \quad 
% & \mathbf{Z}^*_{\text{comb}} = 
% \mathcal{Z}_{\text{argmax}(\mathbf{y})} \\
% & \mathbf{y} = \text{GNN-ReCo}(\mathcal{G}, \mathbf{X}_{\text{cont}}) \\
& f_{\text{MU}} \left(\mathbf{Z}_{\text{comb}}, \mathbf{X}_{\text{cont}} \right) = \begin{cases} 
     m_{\text{prop}},~\text{if } I_{\text{CQI}}^* \leq 2.5 \text{ and } N_{L} = 12 &\\
    \ln\left( (I_{\text{CQI}}^*-2.5)^2+1\right) &\\\quad+
    \ln \left((N_{\text{L}}-12)^2+1\right) + \beta,~\text{otherwise} \\
  \end{cases} \\
& \mathbf{Z}_{\text{comb}} = \mathbf{Z}^{(x_{\text{int}})}\\
& \mathbf{X}_{\text{cont}} =
\begin{bmatrix}
\begin{aligned}
& \Delta x_{1} \dots \Delta x_{4},\ \Delta y_{1} \dots \Delta y_{4}, \\
& v_{1} \dots v_{4},\ m_{\text{MU}},\ r_{\text{thread}},\ r_{\text{corner}},\ L_{\text{net}},\ L_{\text{ct}}
\end{aligned}
\end{bmatrix}\\
\end{aligned}
\end{equation}

\section{Results and Discussion}\label{sec:results}
The training and testing of the \ac{GNN}-ReCo, and the optimization are performed on a dual-socket Intel Xeon Gold 6148 system (40 physical cores, 80 threads) with 187 GB RAM and a 16 GB VRAM NVIDIA GeForce RTX 4060 Ti GPU. In this section, we present an analysis of the training performance of GNN recommender, followed by discussion of the optimization history and results, and its comparison to direct optimization applied to the MCNLP problem as is (without a GNN). The section ends with physical insights into the optimized designs, and their simulated capture performance.

\subsection{Training of the \ac{GNN} Recommender}
The current \ac{GNN} architecture utilizes a 3-layer \ac{GNN} backbone with a hidden dimension of 128. The embeddings are scaled to 64 dimensions to provide richer feature representations. To map graph features to cost predictions, the model employs a deep, 5-layer residual Multi-Layer Perceptron (MLP). Finally, to prevent overfitting given the increased capacity, a moderate dropout rate of 30\% is applied throughout the network. For data collection, 3000 function evaluations are sampled from the high-fidelity simulator, and 100 subgraphs are constructed, each with 30 nodes representing combinatorial features and one common continuous variable. For all simulations, the target is initialized at a position of $\textbf{r}_t = 9\mathbf{\hat{i}}+9\mathbf{\hat{j}}-60\mathbf{\hat{k}}$ m. Its initial orientation is specified by a 3-2-1 Euler angles set (i.e., the set of rotations describing the orientation of the $\mathcal{D}$ frame with respect to the $\mathcal{O}$ frame) of $\{60.0^\circ, 40.0^\circ, 0.0^\circ\}$, while its initial angular velocity (expressed in the $\mathcal{D}$ frame) is defined as $ \boldsymbol{\omega}_t = 10\mathbf{\hat{d}}_x+30\mathbf{\hat{d}}_y+10\mathbf{\hat{d}}_z$ deg/s. This described target condition is considered one of the most challenging target-state scenarios in our previous research~\cite{boonrath2024learning, liu2025surrogate}, as it has both the greatest initial distance from the chaser and the highest angular velocity.

% \begin{figure}[ht!]
%   \centering
%   \includegraphics[width=1.0\linewidth]{Images/per_graph_score_range.pdf}
%   \caption{Shows the training data used to train the GNN-NavCo model. The blue line denotes the min-max range of fitness values each node had and the red dot shows the mean of fitness value}
%    \label{fig:score_per_graph}
% \end{figure}

\begin{figure}[ht!]
  \centering
  \includegraphics[width=1.0\linewidth]{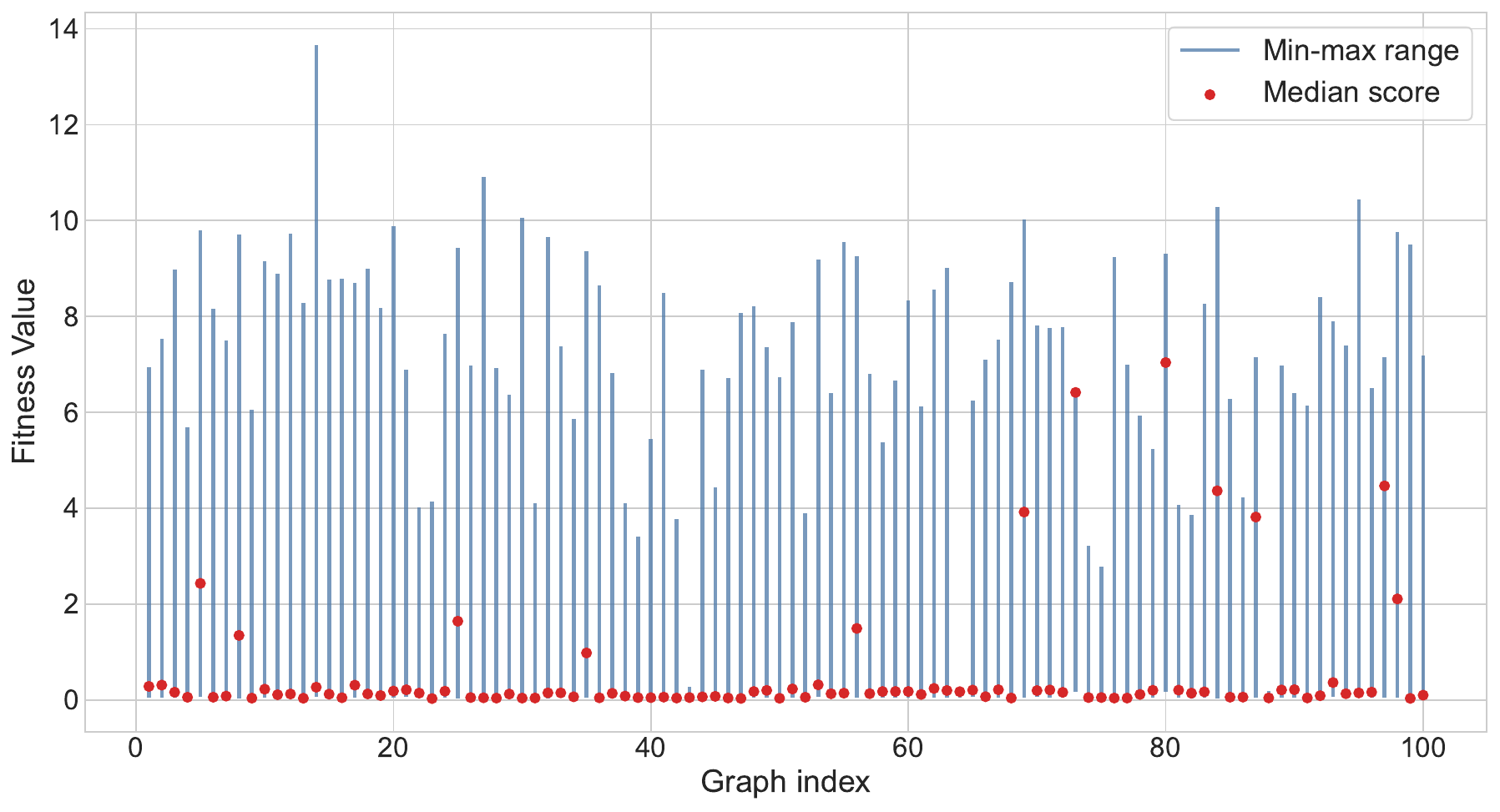}
  \caption{Shows the performance on training data used to train the GNN-NavCo model. The blue line denotes the min-max range of fitness values each node had and the red dot shows the median of fitness value}
   \label{fig:score_per_graph}
\end{figure}

In Fig.~\ref{fig:score_per_graph}, it can be observed that the fitness score $f_{\text{MU}}$ of the collected data varies from near 0 to below 14, which indicates the data covers various successful and failed capture scenarios. By definition of $f_{\text{MU}}$ in Eq.~\ref{eq:obj_MU}, the maximum value for a successful capture will be 0.96. It can be observed that the median fitness values of most of the subgraphs did not exceed the threshold, indicating the objective values were strongly right-skewed. But individual subgraphs still exhibit substantial within-graph variation, as shown by the min-max ranges and mean values in the figure. This variation is useful for training because it exposes the network to both relatively low and high fitness value regions within the same graph. The successful capture rate in the 3,000 samples is 74.5\%. Therefore, it is still necessary to apply the optimization to ensure the capture is successful.

% During data sampling, the objective values were strongly right-skewed, with most samples concentrated below 2. But individual subgraphs still exhibit substantial within-graph variation, as shown by the min-max ranges and mean values in Fig. \ref{fig:score_per_graph}. This variation is useful for training because it exposes the network to both relatively low and high fitness value regions within the same graph. 

%This setup is selected as the scenario with the greatest initial target distance from the chaser with the highest angular velocity, which is considered one of the most challenging target-state scenarios in our previous research~\cite{boonrath2024learning, liu2025surrogate}.

With the collected data, the \ac{GNN}-NavCo model is trained while tracking edge and cycle-consistency regularization losses. Model selection is performed using the minimum validation edge loss, and the corresponding checkpoint is saved at 581 epochs. In Fig.~\ref{fig:comb_hist}, it is observed that the edge loss remains stagnant early on, which can be attributed to the model's early-stage learning of the hidden representation and to the stabilizing effect of the cycle consistency regularization. When edge loss and cycle loss in Fig.~\ref{fig:comb_hist} are compared, it is observed that up until 50 epochs, the cycle loss reduces rapidly. After this point, the edge loss drops sharply, whereas the cycle loss largely plateaus with occasional spikes, although the overall spike is negligible. A similar edge-loss pattern is also observed across experiments with different values of the cycle-loss weight $\lambda_{cycle}$. The abrupt improvement in edge loss is not solely due to the regularizer, but rather to the model learning a representation for edge prediction. This can be observed in Fig.~\ref{fig:comb_hist}, where it is seen that before roughly epochs 50-60, the model has fluctuating sign accuracy, and after the stabilization mark (green line), loss drops and the sign accuracy stabilizes at 75-76\%. Sign accuracy measures whether the model correctly predicts the direction of an edge relation. It can be inferred that early on, the model prioritizes both cyclical consistency and accurate sign prediction, and once it stabilizes, it shifts its focus to reducing edge loss. A hyperparameter study, as shown in Table~\ref{tab:cycle_consistency} over $\lambda_{cycle}$, shows that sign accuracy was broadly similar across the tested values, with $\lambda_{cycle}=0.003$ giving a slightly better overall result; therefore, $\lambda_{cycle}=0.003$ was selected for subsequent experiments. Our parallel research suggests that cycle-consistency regularization can improve generalization, especially in challenging settings such as distribution shift or skewed targets. In the present study, however, the effect of $\lambda_{cycle}$ was modest, with similar sign accuracy across the tested values. In summary, the trained \ac{GNN} can serve as an \textit{instantaneous} recommender because of the negligible execution time of the neural network. Given a continuous vector and a starting combination, the \ac{GNN} has a general 75\% accuracy in recommending a more optimal combination for the continuous vector. The accuracy may seem low, but during the optimization framework, the optimal design is cached thus the actual achieved performance of the \ac{GNN}-aided optimization is not limited much by this accuracy. The accuracy can also be further improved by future research on the \ac{GNN} structure, loss function, or subgraph sampling method, but for the current paper, this trained \ac{GNN} proves to work as Section~\ref{ssec:opt} shows. 

% In Fig.~\ref{fig:score_per_graph}, it is seen that the fitness score values of the dataset is globally concentrated in a relatively low fitness score region, but the individual subgraphs exhibit a substantial variation as evidenced by the min max ranges, while the red dots representing the mean are lying low in the figure, so the cycle consistency regularizer plays a local role towards creating a non zero circulation around the cycles in the graph. In this experiment, $\lambda_{cycle} = 0.003$ is set to apply soft regularization, that is, it improves generalization slightly, as evidenced by the improved validation performance of models trained with cycle regularization in Table~\ref{tab:cycle_consistency}. This shows that, in problems where sampling produces a widely out-of-distribution dataset, using a cyclical regularizer is likely to push the model towards lower validation edge loss and thus better prediction in unseen scenarios. It is observed that the cyclical regularizer acts as an inductive bias toward gradient-like, integrable edge flow fields and away from cyclic or rotational ones.

\begin{figure}[ht!]
  \centering
  \includegraphics[width=0.9\linewidth]{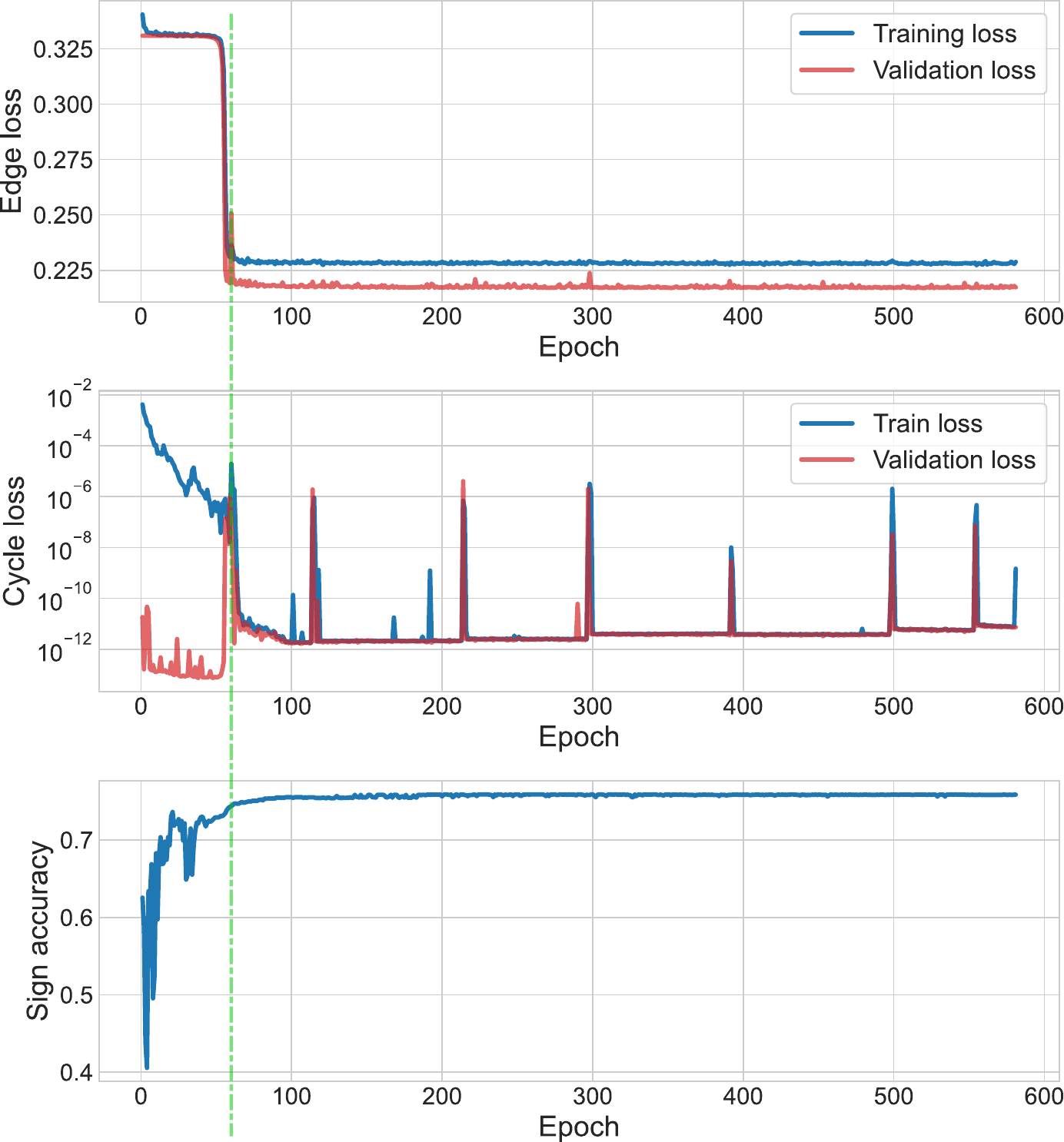}
  \caption{The top figure shows the training and validation loss of the edge loss function, followed by the cycle loss, and the third one shows the sign accuracy curve of the trained model. The green line shows the stabilization point. }
  \label{fig:comb_hist}
\end{figure}

% \begin{figure}[ht!]
%   \centering
%   \includegraphics[width=0.8\linewidth]{Images/edge_loss_train_val.pdf}
%   \caption{Training and validation loss of edge loss function}
%   \label{fig:edge_loss}
% \end{figure}

% \begin{figure}[ht!]
%   \centering
%   \includegraphics[width=0.8\linewidth]{Images/cycle_loss_train_val.pdf}
%   \caption{Training and validation loss of cyclical loss regularizer}
%   \label{fig:cycle_loss}
% \end{figure}

% \begin{figure}[ht!]
%   \centering
%   \includegraphics[width=0.8\linewidth]{Images/sign_accuracy.pdf}
%   \caption{Validation Sign Accuracy of \ac{GNN}-NavCo Model}
%   \label{fig:sign_accuracy}
% \end{figure}

\begin{table}[htbp]
  \centering
  \vspace{2mm}
  \begin{tabular}{lccc}
    \toprule
    \textbf{Hyperparameter} & \textbf{Sign Accuracy} & \textbf{Edge Loss} & \textbf{Cycle Loss} \\ 
    \midrule
    $\lambda_{cycle} = 1$   & 0.76333 & 0.04321 & $1.16 \times 10^{-12}$ \\ 
    $\lambda_{cycle} = 0.003$ & \textbf{0.76356}     & 0.04323     & $2.32 \times 10^{-12}$ \\ 
    $\lambda_{cycle} = 0$   & 0.7623      & 0.043224     & $1.85 \times 10^{-12}$ \\ 
    \bottomrule
  \end{tabular}
    \caption{Experiment on cyclical consistency regularization. sign accuracy, edge loss and cycle loss denote the corresponding converged value at least validation loss}
    \label{tab:cycle_consistency}
\end{table}

\subsection{Optimization Performance and Comparisons between with/without GNN recommendation}\label{ssec:opt}
% \section{Results and Discussion}\label{sec:results}
The total computing time to finish 53 iterations of \ac{GNN}-NavCo-aided MDPSO is 26.5 hours. The convergence history of the objective is shown in Fig.~\ref{fig:conv}. The sequential quadratic programming optimizer (implemented by MATLAB \texttt{fmincon()}) is applied on the continuous variables optimized by \ac{GNN}-NavCo-aided MDPSO to further investigate the potential improvement, but failed to find any more optimal results, proving the current result is already a local optimal. The original MDPSO converged at the 27th iteration, while the \ac{GNN}-NavCo-aided converged at the 8th iteration. Here, convergence is defined as the point at which the absolute difference between the current objective value falls into the lower 0.5\% of the total range of objective values observed over the entire optimization history. Though the MDPSO without GNN eventually reached to an objective slightly better than the GNN-aided one, it requires more iterations to reach the optimized objective of the GNN-aided MDPSO, which terminated earlier due to the stopping criteria mentioned in Section~\ref{sec:frame}. In summary, the \ac{GNN}-NavCo used 1,900 fewer function evaluations during the optimization to reach convergence. It needs to be noted that the \textit{total} function evaluations used by the \ac{GNN}-NavCo-aided \ac{MDPSO} are 1,100 more (considering the 3,000 function evaluations used for training the GNN) than those of the \ac{MDPSO} without the aid of \ac{GNN}. However, the \ac{GNN}-NavCo-aided methods remain promising.

First, the additional function evaluations are primarily associated with the data-generation and training phases of the \ac{GNN} surrogate, which can be performed \textit{offline}. This process is highly parallelizable and can be efficiently executed on high-performance computing (HPC) platforms or supercomputers, thereby significantly reducing the computing time despite the increased number of evaluations. In contrast, the baseline \ac{MDPSO} relies on sequential updates between iterations as online evaluations during optimization, which are typically more time-constrained.

Second, from a scalability perspective, the advantage of the \ac{GNN}-NavCo framework is expected to become more pronounced as the size and complexity of the combinatorial space increase. The ability to reduce data-generation costs and leverage learned structural information makes the approach more suitable for large-scale problems, where purely optimization-based methods may struggle to efficiently explore the search space. Our ongoing parallel research will further investigate this statement.
\begin{figure*}[ht!]
  \centering
  \includegraphics[width=0.9\linewidth]{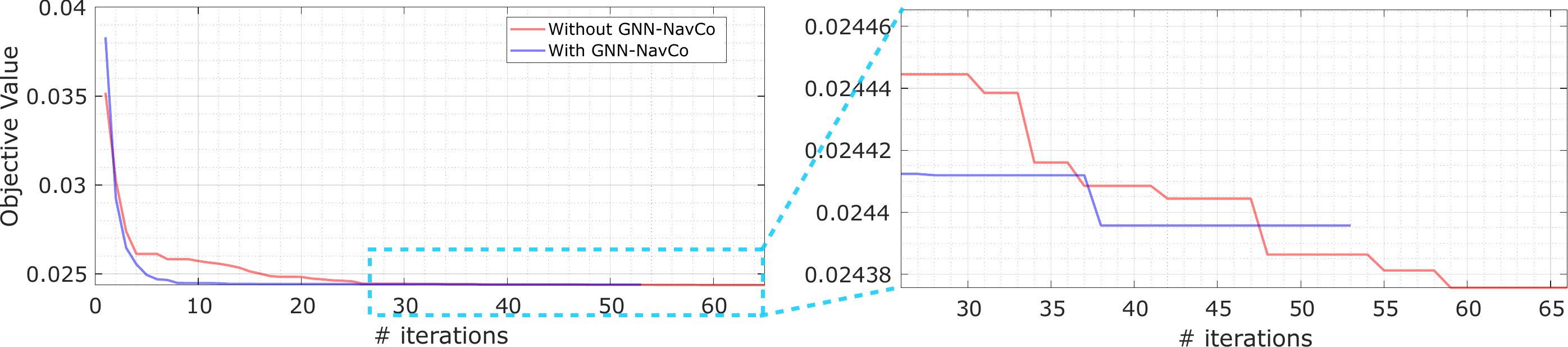}
  \caption{Objective convergence history comparison between the optimization with and without the aid of \ac{GNN}-NavCo}
  \label{fig:conv}
\end{figure*}

\subsection{Physical Insights on the Optimized and Baseline Designs and Their Capture Performance}
The comparison of the design variables and optimized objectives is shown in Table~\ref{tab:comp}. Both optimization methods converged on the same combination: a thruster with the highest $I_{\text{sp}}$ and the 2nd-highest thrust, and a net with the fewest meshes, which together significantly reduce the system's fuel consumption. It should be noted that although the combinatorial variables are identical between the two methods, several continuous variables—such as the aiming points and the net side length—differ significantly. This observation suggests that the underlying optimization problem is highly nonlinear and admits multiple feasible solutions with comparable performance. In other words, even under the same discrete configuration, the continuous design space likely contains multiple local optima or near-optimal regions that can yield similar objective values. The differences observed in the continuous variables indicate that the two methods have converged to distinct solutions within this landscape, rather than a single unique optimum. This highlights the problem's inherent multi-modality, in which different values of the continuous variables can compensate for one another to achieve similar system performance. As a result, the final solution is not necessarily unique, and the optimizer may converge to different, yet equally valid, designs even though they have exactly the same initial populations. The rendered simulation screenshots of the \ac{GNN}-NavCo-aided optimized design are shown in Fig.~\ref{fig:render}. The animated capture of the three methods are provided in a GitHub repository~\cite{githubrepo_aiaa}. Since all three methods achieved successful capture in this scenario, here the $f_{\text{MU}}$ equals the fuel cost. Compared with the baseline design from our previous study \cite{boonrath2024learning}, the optimized design consumes only 14.35\% of the baseline's fuel, highlighting the critical role of optimization (in conjunction with the MPC) in remarkably improving system fuel efficiency.

\begin{table*}[ht]
% \footnotesize
\centering
\caption{Comparison between optimized variables and objective with and without the aid of \ac{GNN}-NavCo}
\begin{tabular}{|l|c|c|c|}
\hline
\textbf{Continuous Variable} & \ac{GNN}-NavCo-aided MDPSO & MDPSO & Baseline\\
\hline
$\Delta x_1, \dots \Delta x_4$ [m] & [2.843, -4.980, 0.041, -4.997] & [-1.132, -4.060, 4.286, -4.999] & [0,0,0,0]\\
% \hline
$\Delta y_1, \dots \Delta y_4$ [m] & [0.569, 4.103, -1.799, -0.880] & [1.794, 1.963, -4.144, -2.404] & [0,0,0,0]\\
% \hline
$v_1, \dots, v_4$ [m/s] & [0.527, 0.785, 0.837, 0.841] & [0.654, 0.733, 0.846, 0.767] & [0,0,0,0]\\
% \hline
$m_{\text{MU}}$ [kg] & 2.1862 & 2.7569 & 2.0 \\
% \hline
$r_{\text{thread}}$ [m] & 0.0011 & 0.0011 & 0.0010\\
$r_{\text{corner}}$ [m] & 0.0012 & 0.0009 & 0.0010 \\
$L_{\text{net}}$ [m] & 23.315 & 19.667 & 22.0\\
$L_{\text{ct}}$ [m] & 1.532 & 0.931 & 2.414\\
\hline
\textbf{Combinatorial Variable} & & &\\
\hline
$F_{T,\text{max}}$ [N] & 6.1 & 6.1 & 8.9 \\
$I_{\text{sp}}$ [s] & 277.0 & 277.0 & 60.0 \\
$m_{T}$ [kg] & 0.6 & 0.6 & 0.37\\
$E_{\text{n}}$ [Gpa] & 70.0 & 70.0 & 70.0 \\
$\rho_{\text{n}}$ [kg/m$^3$] & 1390.0 & 1390.0 & 1390.0 \\
$N_k$ & 9 & 9 & 11\\
$K_{\text{cls}}$ & 0 & 0 & 1 \\
\hline
\textbf{Objective} & & & \\
\hline
$f_{\text{MU}}$ & 0.0244 & 0.0244 & 0.170\\
\hline

\end{tabular}
\label{tab:comp}
% \vspace{-5pt}
\end{table*}

\begin{figure*}[ht!]
  \centering
  \includegraphics[width=1.0\linewidth]{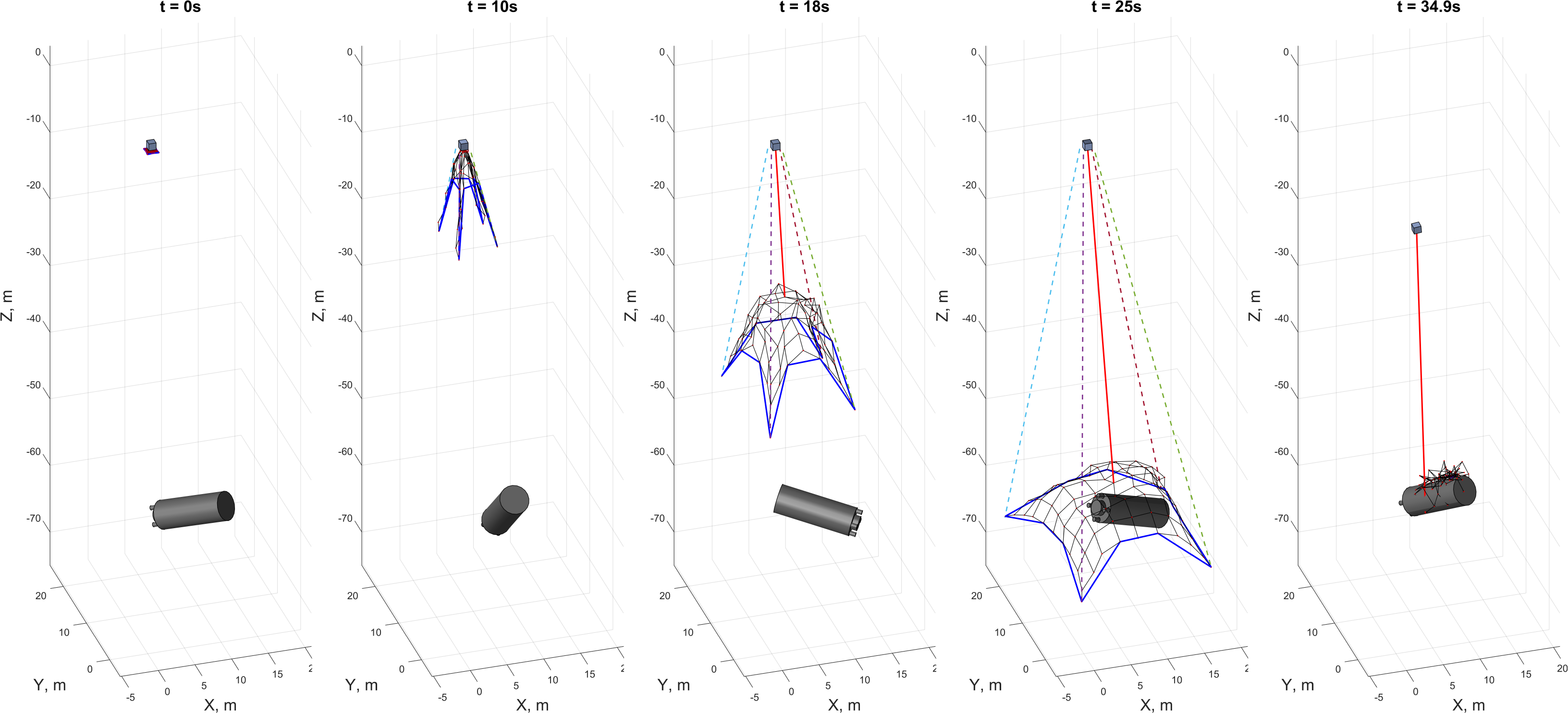}
  \caption{Rendered simulation screenshots of the \ac{GNN}-NavCo-aided optimized design. The dashed lines are the minimum-energy reference trajectories for the \acp{MU}. The following frames are shown in the above figure. \textbf{1) t=0s:} The beginning of the entire mission. \textbf{2) t=10s and 18s:} The MUs are maneuvered towards the optimized aiming points. \textbf{3) t=25s:} The thrusters are switched off, and the closing mechanism is switched on. \textbf{4) t=34.9s:} The capture is stabilized.}
  \label{fig:render}
\end{figure*}

\section{Conclusion}\label{sec:con}
In this paper, we developed a \ac{GNN}-aided nonlinear optimization framework to search the relatively high-dimensional mixed-combinatorial space presented by joint consideration of the physical design of the net, component choices and aiming point control decisions for a tether-net system used in space debris capture. The presented framework found a converged physical design and control that provided a 7$\times$ reduction in fuel costs of the tether-net maneuver compared to a baseline design from previously reported work, while maintaining successful capture. While both optimizations were allowed to run for a bit more than 50 iterations using the PSO algorithm (post-processed by SQP), the new GNN-aided PSO approach achieved the threshold convergence tolerance of within 0.0001 kg of the optimum fuel cost in only 800 function evaluations, compared to the 2700 function evaluations taken by the standard PSO (that solves the entire MCNLP without GNN) to achieve the same. In terms of compute time, this was a savings of 9.5 hours, relative to overall run time in the order of about 25 hours. 

It is also important to note that the integration of the \ac{GNN}-NavCo module introduces a learning-based guidance mechanism that captures structural relationships within the combinatorial design space, enabling more informed exploration compared to purely stochastic search. Sample generation and training the GNN is therefore a one-time overhead, which can then be used with multiple different optimization runs with different settings. This computational overhead can also be largely amortized through offline processing and parallel execution on high-performance computing platforms. 
% Once trained, the GNN provides rapid inference that effectively steers the optimizer toward promising regions of the search space, thereby improving convergence behavior. This suggests that the proposed framework is particularly promising for larger-scale and more complex mixed-combinatorial problems, where traditional methods may struggle to efficiently navigate the search space.

In its current form, the design space does not include sensor choices or a variety of net shapes, addition of which would provide more meaningful exploration of the design space for autonomous tether-net systems. In addition, application of the \ac{GNN}-NavCo module with other optimization algorithms, especially gradient based methods and Bayesian optimization, with the potential to update the GNN insitu during optimization could provide further insights into the generalized applicability of this approach for automated design guidance of mixed-combinatorial complex systems. 
% Additionally, our previous research has shown that the controls of the tether-net system strongly influence capture results. Therefore, the control variables of the tether-net can be fine-tuned using reinforcement-learning-based methods, leveraging the optimized physical design presented in this manuscript.

% \section*{Appendix}

% An Appendix, if needed, should appear before the acknowledgments.

\section*{Acknowledgments}
This work is supported under the CMMI Awards numbered 2128578 and 2048020 from the National Science Foundation (NSF). The authors' opinions, findings, and conclusions or recommendations expressed in this material do not necessarily reflect the views of the National Science Foundation. The University at Buffalo Presidential Fellowship is acknowledged by Achira Boonrath.

\bibliographystyle{IEEEtran}
\bibliography{sample} % Path to your .bib file

@article{paul2024learning,
    author = {Paul, Steve and Chowdhury, Souma},
    title = {Learning to Allocate Time-Bound and Dynamic Tasks to Multiple Robots Using Covariant Attention Neural Networks},
    journal = {Journal of Computing and Information Science in Engineering},
    volume = {24},
    number = {9},
    pages = {091005},
    year = {2024},
    month = {08},
    doi={10.1115/1.4065883}, 
    issn = {1530-9827},
}

@article{vaswani2017attention,
  title={Attention is all you need},
  author={Vaswani, Ashish and Shazeer, Noam and Parmar, Niki and Uszkoreit, Jakob and Jones, Llion and Gomez, Aidan N and Kaiser, {\L}ukasz and Polosukhin, Illia},
  journal={Advances in neural information processing systems},
  volume={30},
  year={2017}
}

@article{tong2016multi,
  title={A multi-objective mixed-discrete particle swarm optimization with multi-domain diversity preservation},
  author={Tong, Weiyang and Chowdhury, Souma and Messac, Achille},
  journal={Structural and Multidisciplinary Optimization},
  volume={53},
  pages={471--488},
  year={2016},
  doi={10.1007/s00158-015-1319-8},
  publisher={Springer}
}

@article{chowdhury2013mixed,
  title={A mixed-discrete particle swarm optimization algorithm with explicit diversity-preservation},
  author={Chowdhury, Souma and Tong, Weiyang and Messac, Achille and Zhang, Jie},
  journal={Structural and Multidisciplinary Optimization},
  volume={47},
  pages={367--388},
  year={2013},
  doi={10.1007/s00158-012-0851-z},
  publisher={Springer}
}

@article{ZhangShiqiang2024Aomd,
author = {Zhang, Shiqiang and Campos, Juan S. and Feldmann, Christian and Sandfort, Frederik and Mathea, Miriam and Misener, Ruth},
copyright = {2024 The Author(s)},
issn = {0098-1354},
journal = {Computers \& chemical engineering},
language = {eng},
pages = {108684-},
publisher = {Elsevier Ltd},
title = {Augmenting optimization-based molecular design with graph neural networks},
doi={10.1016/j.compchemeng.2024.108684},
volume = {186},
year = {2024},
}

@article{XIA2025103415,
title = {Mixed integer programming modeling for the satellite three-dimensional component assignment and layout optimization problem},
journal = {Chinese Journal of Aeronautics},
pages = {103415},
year = {2025},
issn = {1000-9361},
author = {Yufeng Xia and Xianqi Chen and Zhijia Liu and Weien Zhou and Wen Yao and Zhongneng Zhang},
doi= {10.1016/j.cja.2025.103415},
}

@INPROCEEDINGS{PSO,
  author={Kennedy, J. and Eberhart, R.},
  booktitle={Proceedings of ICNN'95 - International Conference on Neural Networks}, 
  title={Particle swarm optimization}, 
  year={1995},
  volume={4},
  number={},
  pages={1942-1948 vol.4},
}

@article{RingkampMaik2017Ottt,
author = {Ringkamp, Maik and Ober-Blöbaum, Sina and Leyendecker, Sigrid},
address = {Berlin/Heidelberg},
copyright = {Springer-Verlag Berlin Heidelberg and Mathematical Optimization Society 2016},
issn = {0025-5610},
journal = {Mathematical programming},
language = {eng},
number = {1-2},
pages = {551-581},
publisher = {Springer Berlin Heidelberg},
title = {On the time transformation of mixed integer optimal control problems using a consistent fixed integer control function},
volume = {161},
doi={10.1007/s10107-016-1023-5},
year = {2017},
}

@article{jang2023active,
  title={Active Debris Removal Simulations Using Spider-Web Space-Nets for KITSAT-1 Satellite},
  author={Jang, Mi and Shin, Hyun-Cheol and Sim, Chang-Hoon and Park, Jae-Sang and Cho, Haeseong},
  journal={International Journal of Aeronautical and Space Sciences},
  volume={24},
  number={5},
  doi={10.1007/s42405-023-00623-2},
  pages={1311--1322},
  year={2023},
  publisher={Springer}
}

@article{ref2,
title = {Solve routing problems with a residual edge-graph attention neural network},
journal = {Neurocomputing},
volume = {508},
pages = {79-98},
year = {2022},
issn = {0925-2312},
doi={10.1016/j.neucom.2022.08.005},
author = {Kun Lei and Peng Guo and Yi Wang and Xiao Wu and Wenchao Zhao}
}

@INPROCEEDINGS{ref3,
  author={Paul, Steve and Ghassemi, Payam and Chowdhury, Souma},
  booktitle={2022 International Conference on Robotics and Automation (ICRA)}, 
  title={Learning Scalable Policies over Graphs for Multi-Robot Task Allocation using Capsule Attention Networks}, 
  year={2022},
  volume={},
  number={},
  pages={8815-8822},
  doi={10.1109/ICRA46639.2022.9812370},
}

@Article{ref5,
author={Zhang, Si
and Tong, Hanghang
and Xu, Jiejun
and Maciejewski, Ross},
title={Graph convolutional networks: a comprehensive review},
journal={Computational Social Networks},
year={2019},
month={Nov},
day={10},
doi={10.1186/s40649-019-0069-y},
volume={6},
number={1},
pages={11},
issn={2197-4314},
}

@inproceedings{
ref6,
title={Graph Attention Networks},
author={Petar Veličković and Guillem Cucurull and Arantxa Casanova and Adriana Romero and Pietro Liò and Yoshua Bengio},
booktitle={International Conference on Learning Representations},
  volume={6},
  number={2},
  year={2018},
  doi = {},
  organization={International Conference on Learning Representations}
}

@inproceedings{ref7,
      title={Graph Capsule Convolutional Neural Networks}, 
      author={Saurabh Verma and Zhi-Li Zhang},
      year={2018},
      booktitle={arXiv},
      primaryClass={stat.ML},
}

@misc{ecaps_brochure_2023,
  author       = {ECAPS},
  title        = {ECAPS 5N HPGP Thruster Brochure},
  year         = 2023,
  url          = {https://satsearch.co/products/ecaps-5n-hpgp-thruster},
  note         = {Accessed: 2025-03-20},
}

@misc{rubicon_5n_lt_thruster,
  author       = {Rubicon Space Systems},
  title        = {5N Low Throughput (LT) Thruster},
  year         = 2023,
  url          = {https://www.rubicon.space/product/19/5n-lt-thruster},
  note         = {Accessed: 2025-03-20},
}

@manual{kevlar_technical_guide,
  author       = {DuPont},
  title        = {Kevlar® Aramid Fiber Technical Guide},
  year         = 2023,
  url          = {https://www.dupont.com/products/kevlar},
  note         = {Accessed: 2025-03-20},
}

@techreport{nasa2025soa,
  author = {Sasha V. Weston and Craig D. Burkhard and Jan M. Stupl and Rachel L. Ticknor and Bruce D. Yost and Rebekah A. Austin and Pavel Galchenko and Lauri K. Newman and Luis Santos Soto},
  title = {State-of-the-Art Small Spacecraft Technology 2024},
  institution = {National Aeronautics and Space Administration (NASA)},
  year = {2025},
  month = {February},
  number = {NASA/TP—20250000142},
  address = {Ames Research Center, Moffett Field, California},
  url = {https://www.nasa.gov/smallsat-institute/sst-soa}
}

@article{liu2025surrogate,
  title={Surrogate-aided Learning of Active Tether-Net Maneuver to Capture Rotating Space Debris},
  author={Liu, Feng and Boonrath, Achira and Botta, Eleonora M and Chowdhury, Souma},
  journal={IEEE Transactions on Aerospace and Electronic Systems},
  year={2025},
  volume={62},
  pages={630 - 645},
doi={10.1109/TAES.2025.3624187},
  publisher={IEEE}
}

@article{GNN_survey,
  author       = {Zonghan Wu and
                  Shirui Pan and
                  Fengwen Chen and
                  Guodong Long and
                  Chengqi Zhang and
                  Philip S. Yu},
  title        = {A Comprehensive Survey on Graph Neural Networks},
  journal      = {CoRR},
  volume       = {abs/1901.00596},
  year         = {2019},
  eprinttype    = {arXiv},
  doi={10.1109/TNNLS.2020.2978386},
}

@article{GNN_distill,
  author = {Sanchez-Lengeling, Benjamin and Reif, Emily and Pearce, Adam and Wiltschko, Alexander B.},
  title = {A Gentle Introduction to Graph Neural Networks},
  journal = {Distill},
  year = {2021},
  note = {https://distill.pub/2021/gnn-intro},
}

@inproceedings{gnn_reco_idetc,
    title = {Efficient Design Optimization Over Mixed-Combinatorial Spaces Enabled by Graph-Learning},
    author = {Liu, Feng and Chowdhury, Souma and Boonrath, Achira and Botta, Eleonora M.},
    pages = {V03AT03A021},
    booktitle = {ASME 2025 International Design Engineering Technical Conferences and Computers and Information in Engineering Conference},
    year = {2025},
    month = {08},
    doi = {10.1115/DETC2025-169719},
    eprint = {https://asmedigitalcollection.asme.org/IDETC-CIE/proceedings-pdf/IDETC-CIE2025/89220/V03AT03A021/7557501/v03at03a021-detc2025-169719.pdf},
}

@phdthesis{BottaPhdThesis,
  author       = {Eleonora M. Botta}, 
  title        = {Deployment and {Capture} {Dynamics} of {Tether}-{Nets} for {Active} {Space} {Debris} {Removal} },
  school       = {McGill University},
  year         = 2017,
  address      = {Montreal, Québec }
}

@article{boonrath2024learning,
  title={Learning-Aided Control of Robotic Tether-Net with Maneuverable Nodes to Capture Large Space Debris},
  author={Boonrath, Achira and Liu, Feng and Botta, Elenora M and Chowdhury, Souma},
  journal={IEEE 2024 International Conference on Robotics and Automation},
  year={2024},
  doi={10.1109/ICRA57147.2024.10610721}
}

@article{jsr2023Zeng,
author = {Zeng, Chen and Hecht, Grant R. and Chowdhury, Souma and Botta, Eleonora M.},
title = {Concurrent Design Optimization of Tether-Net System and Actions for Reliable Space-Debris Capture},
journal = {Journal of Spacecraft and Rockets},
volume = {0},
number = {0},
pages = {1-11},
year = {0},
doi = {10.2514/1.A35812},
URL = {https://doi.org/10.2514/1.A35812 },
}

@article{boonrath2025mission,
  title={A mission concept for removing multiple debris using formation-flying slingshot spacecraft},
  author={Boonrath, Achira and Rossi, Federico and Nesnas, Issa A and Botta, Eleonora M},
  journal={Acta Astronautica},
  year={2025},
  volume = {237},
pages = {443-459},
doi={10.1016/j.actaastro.2025.08.035},
  publisher={Elsevier}
}

@article{ZhaoYakun2020ISSM,
pages = {6874-6882},
author = {Zhao, Yakun and Zhang, Fan and Huang, Panfeng and Liu, Xiyao},
issn = {0278-0046},
journal = {IEEE transactions on industrial electronics (1982)},
language = {eng},
number = {8},
publisher = {IEEE},
title = {Impulsive Super-Twisting Sliding Mode Control for Space Debris Capturing via Tethered Space Net Robot},
volume = {67},
doi={10.1109/TIE.2019.2940002},
year = {2020},
}

@article{botta2016simulation,
  title={On the simulation of tether-nets for space debris capture with {Vortex} {Dynamics} },
  author={Botta, Eleonora M and Sharf, Inna and Misra, Arun K and Teichmann, Marek},
  journal={Acta Astronautica},
  volume={123},
  pages={91--102},
  year={2016},
  publisher={Elsevier},
  doi={10.1016/j.actaastro.2016.02.012}
}

@article{botta2019simulation,
  title={Simulation of tether-nets for capture of space debris and small asteroids},
  author={Botta, Eleonora M and Sharf, Inna and Misra, Arun K},
  journal={Acta Astronautica},
  volume={155},
  pages={448--461},
  year={2019},
  publisher={Elsevier},
  doi={10.1016/j.actaastro.2018.07.046}
}

@article{hou2021dynamic,
  title={Dynamic computation of a tether-net system capturing a space target via discrete elastic rods and an energy-conserving integrator},
  author={Hou, Yunsen and Liu, Cheng and Hu, Haiyan and Yang, Wenmiao and Shi, Junwei},
  journal={Acta Astronautica},
  volume={186},
  pages={118--134},
  year={2021},
doi = {10.1016/j.actaastro.2021.05.029},
publisher={Elsevier}
}

@article{rl,
  author    = {Jonathan Baxter and
               Andrew Tridgell and
               Lex Weaver},
  title     = {KnightCap: {A} chess program that learns by combining TD(lambda) with
               game-tree search},
  journal   = {CoRR},
  volume    = {cs.LG/9901002},
  year      = {1999},
  url       = {https://arxiv.org/abs/cs/9901002},
  bibsource = {dblp computer science bibliography, https://dblp.org}
}

@article{cqi,
author = {Barnes, Charles and Botta, Eleonora},
year = {2020},
month = {07},
pages = {455-463},
title = {A quality index for net-based capture of space debris},
volume = {176},
journal = {Acta Astronautica},
doi = {10.1016/j.actaastro.2020.06.044}
}

@inproceedings{anzmeadorIridiumCosmos,
  title={Analysis and {Consequences} of the {Iridium} 33-{Cosmos} 2251 {Collision} },
  author={Anz-Meador, Phillip and Liou, Jer-Chyi},
  booktitle={38th COSPAR Scientific Assembly},
  year={July 2010},
  address = {Bremen, Germany}
}

@article{boonrath2025robustness,
  title={Robustness and Safety of Net-Based Debris Capture Under Deployment and Environmental Uncertainties},
  author={Boonrath, Achira and Botta, Eleonora M},
  journal={Journal of Spacecraft and Rockets},
  pages={1--17},
  year={2025},
doi={10.2514/1.A36217},
  publisher={American Institute of Aeronautics and Astronautics}
}

@inproceedings{liu2023learning,
  title={Learning Constrained Corner Node Trajectories of a Tether Net System for Space Debris Capture},
  author={Liu, Feng and Boonrath, Achira and KrisshnaKumar, Prajit and Botta, Eleonora M and Chowdhury, Souma},
  booktitle={AIAA AVIATION 2023 Forum},
doi={10.2514/6.2023-3920},
  pages={3920},
  year={2023}
}

@misc{githubrepo_aiaa,
  author = {Liu, Feng and Boonrath, Achira and Madhu, Gishnu and Botta, Eleonora M. and Chowdhury, Souma},
  title = {Data and video repository for the paper -- Designing Active Tether-Net System for Space Debris Capture with Graph Learning aided Mixed-Combinatorial Optimization},
  year = {2026},
  howpublished = {\url{https://github.com/adamslab-ub/GNN-ReCo-Benchmark/tree/GNN_NavCo}},
}

@article{zhu2024multi,
  title={Multi-Debris Capture by Tethered Space Net Robot via Redeployment and Assembly},
  author={Zhu, Weiliang and Pang, Zhaojun and Du, Zhonghua and Gao, Guangfa and Zhu, Zheng H},
  journal={Journal of Guidance, Control, and Dynamics},
  pages={1--18},
  year={2024},
  doi={10.2514/1.G007908},
  publisher={American Institute of Aeronautics and Astronautics}
}

@article{huang2023contact,
  title={Contact dynamic analysis of tether-net system for space debris capture using incremental potential formulation},
  author={Huang, Weicheng and Zou, Huaiwu and Liu, Hanwu and Yang, Wenmiao and Gao, Jinling and Liu, Zhaowei},
  journal={Advances in Space Research},
doi = {10.1016/j.asr.2023.05.054},
  year={2023},
  publisher={Elsevier}
}

@ARTICLE{approach,
  author={Meng, Zhongjie and Huang, Panfeng and Guo, Jian},
  journal={IEEE Transactions on Aerospace and Electronic Systems}, 
  title={Approach Modeling and Control of an Autonomous Maneuverable Space Net}, 
  year={2017},
  volume={53},
  number={6},
  pages={2651-2661},
  doi={10.1109/TAES.2017.2709794}
}

@article{shanADRreview,
  title={A Review and comparison of active space debris capturing and removal methods},
  author={Shan, Minghe and Guo, Jian and Gill, Eberhard},
 journal={Progress in Aerospace Sciences},
  volume={80},
  pages={18--32},
  year={2015},
  publisher={Elsevier},
  doi={10.1016/j.paerosci.2015.11.001}
}

@INPROCEEDINGS{leastsquares,
  author={Hirani, Anil N. and Kalyanaraman, Kaushik and Watts, Seth},
  booktitle={2015 IEEE International Parallel and Distributed Processing Symposium Workshop}, 
  title={Graph Laplacians and Least Squares on Graphs}, 
  year={2015},
  volume={},
  number={},
  pages={812-821},
  doi={10.1109/IPDPSW.2015.73}}
% \footnote*{This work was presented at the AIAA Aviation 2023 Forum}

\end{document}